\documentclass[lettersize,journal]{IEEEtran}
\ifCLASSINFOpdf
\usepackage[utf8]{inputenc} 
\usepackage[T1]{fontenc}    
\usepackage[hidelinks]{hyperref}       
\usepackage{url}            
\usepackage{booktabs}       
\usepackage{amsfonts}       
\usepackage{nicefrac}      
\usepackage{microtype}      
\usepackage{xcolor}        
\usepackage[numbers]{natbib}  
\usepackage{bm}
\usepackage{amsmath}
\usepackage{amssymb}
\usepackage{multirow}
\usepackage{subcaption}              
\usepackage{array}
\usepackage{amsthm}
\usepackage{rotating}
\usepackage{pifont}
\usepackage{wrapfig}
\usepackage{caption} 
\usepackage{algorithm}      
\usepackage{algorithmicx}  
\usepackage{algpseudocode}  
\usepackage{graphicx}
\usepackage{caption}
\usepackage{graphicx}
\usepackage{graphics}

\begin{document}
	\title{CoDCL: Counterfactual-Inspired Augmentation Contrastive Learning for Temporal Link Prediction in Social Networks}
	\author{Hantong Feng, Duxin Chen, \textit{Member, IEEE}, Wenwu Yu, \textit{Senior Member, IEEE}
		
		\thanks{
			
			Hantong Feng is with the Jiangsu Key Laboratory of Net-Worked Collective Intelligence, School of Cyber Science and Engineering,
			Southeast University, Nanjing 210096, China.
			
			Duxin Chen is with the School of Mathematics, Southeast University,
			Nanjing 210096, China.
			
			Wenwu Yu is with the Frontiers Science Center for Mobile Information
			Communication and Security, School of Mathematics, Southeast University,
			Nanjing 210096, China, and also with the Purple Mountain Laboratories,
			Nanjing 211102, China.
	}}
	\maketitle
	\maketitle
	\maketitle
	\begin{abstract}
		Temporal link prediction is crucial for rapidly growing social networks. Existing methods often overlook the underlying causal mechanisms that drive link formation, making it difficult for algorithms to adapt to complex structures that continuously evolve over time. To enable prediction models to adapt to complex temporal environments, they need to be robust to emerging structural changes. We propose a dynamic network learning framework CoDCL, which combines counterfactual-inspired augmentation with contrastive learning to address this deficiency.
		Furthermore, we devise a comprehensive strategy to generate high-quality counterfactual data, combining a dynamic treatments design with efficient structural neighborhood exploration to quantify the temporal changes in interaction patterns.
		Crucially, the entire CoDCL is designed as a plug-and-play universal module that can be seamlessly integrated into various existing temporal graph models without requiring architectural modifications.
		Extensive experiments conducted on multiple real-world datasets demonstrate that CoDCL significantly outperforms state-of-the-art baselines in temporal link prediction, highlighting the effectiveness of integrating counterfactual-inspired data augmentation into dynamic representation learning.
		
	\end{abstract}
	
	\section{Introduction}
	\IEEEPARstart{T}{emporal} link prediction has become a fundamental challenge in dynamic social network learning \cite{11202740,10930952}.  
	Unlike static networks, temporal networks present unique challenges: relationships evolve continuously over time, interaction timing carries critical information, and historical patterns significantly influence future link formation \cite{10705800}. With the deepening of understanding of time series network, research in continuous-time dynamic network link prediction has emerged to specifically mine the rich spatiotemporal information embedded in dynamic graph structures for future link prediction. 
	However, designing prediction algorithms that can adapt to complex and variable scenarios remains an open and challenging research problem. This is especially true in network problems, because newly emerging nodes and edges may be completely unrelated to the training data and may lack possible simulations of their evolutionary states~\cite{guidotti2024counterfactual,melistas2024benchmarking}.
	
	Most current temporal link prediction methods employ black-box approaches \cite{11027145}. In this case, the most efficient way to achieve accurate prediction of temporal links is through an inductive generalization representation. These representations can be learned directly from known or learned abstract concepts, primarily through complex algorithms to learn local structures, or by evolving network dynamics through historical temporal informatione~\cite{10475162,wang2025dynamic}. We extend previous work by attempting to derive induction from causal knowledge present in the context. Researchers have already attempted to explore the true causality of networks in similar static network domains. However, dynamic social networks, due to their temporal and dynamic nature, inevitably face more challenges, mainly in two aspects~\cite{zhao2022learning}. First, factual explanations cannot explore how input variations lead to different prediction results; they can only identify statistical correlations but cannot distinguish between true causal relationships and spurious correlations \cite{10266657}. Users find it difficult to understand which factors truly play a decisive role and which are merely coincidental co-occurrence patterns, thus hindering performance enhancements for link prediction~\cite{11037524,xie2023factual}.
	
	Secondly, due to the constantly changing patterns between the test environment and the training environment in dynamic network scenarios, previously significant historical events may lose their predictive value at a slightly later stage. Importantly, how to establish a dynamic temporal causal relationship of "temporal evolution pattern - future link" to enhance link prediction is also a pressing problem we need to solve.
	
	Our key insight is that prediction algorithms can improve the model's generalization ability to temporal changes by accurately capturing local causality. Consider a dynamic social network where \textit{Tom} and \textit{Jerry} frequently interact and share neighbors. Traditional methods can learn the strong association between temporal co-occurrence patterns and friendship formation, but may fail to capture the true underlying causes, as shown in Fig.1. To address this, we construct counterfactual samples: while maintaining the basic local structure, we perturb the temporal distribution of their interactions and compare it with the original samples to learn constraints, making the model more robust to new temporal changes.
	\begin{figure}[h]
		\centering
		\includegraphics[scale=0.4]{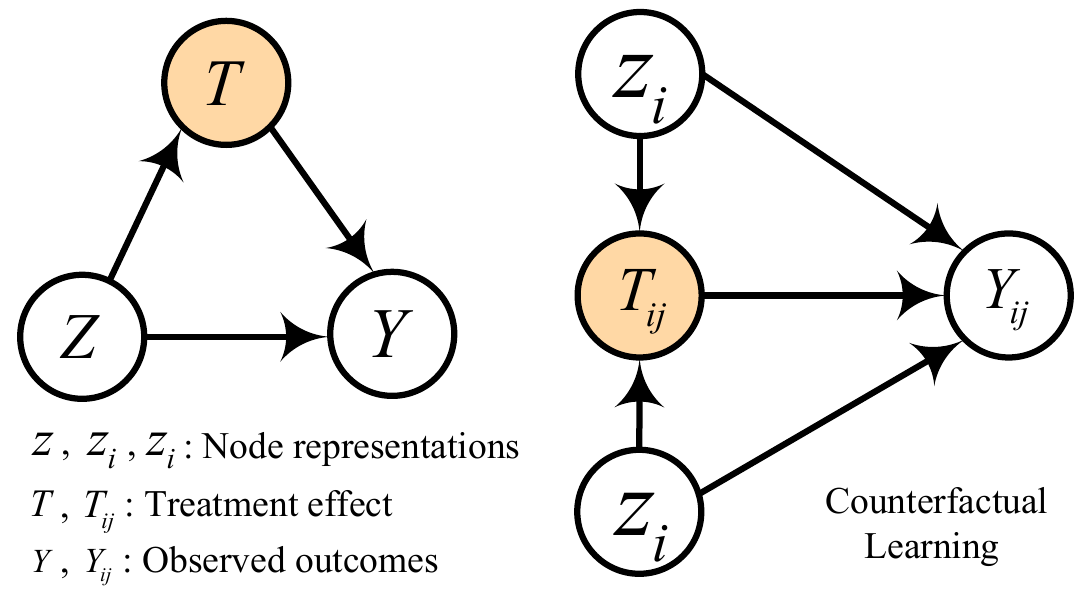}
		\caption{A simple toy example of counterfactual-enhanced link prediction. Causal modeling: Given $Z$ and observed outcomes, find treatment effect of $T$ on $Y$ (Left); Counterfactual model: leverage the estimated treatment effect $(A_{i,j} |T_{i,j})$ to improve the learning of $z_i$ and $z_j$ (Right).}
		\label{4}
	\end{figure}
	
	To bridge this gap, We introduce Counterfactual-inspired augmentation Dynamic network Contrastive Learning(CoDCL), a novel framework that generates counterfactual links and learns from both factual and counterfactual temporal patterns. 
	Specifically, for each observed node pair, we identify the most similar pair sets in feature space through neighbor breadth search, create counterfactual links based on dynamic interaction metrics, and learn representations from observed and counterfactual temporal patterns through contrastive learning. This encourages the model to learn representations that capture essential link formation factors beyond superficial temporal associations. 
	Our contributions are as follows:
	\begin{itemize}
		\item Our proposed algorithm CoDCL innovatively performs data augmentation in local dynamic networks based on counterfactual causality and enhances the temporal network representation from observed and counterfactual temporal patterns through contrastive learning.
		
		\item CoDCL overcomes the complex variations of temporal networks by combining dynamic treatment variables with efficient temporal nearest neighbor search, designing a novel counterfactual pair identification strategy to enhance model interpretability.
		
		\item We validate CoDCL as a plug-and-play enhancement module that can be seamlessly integrated into existing temporal architectures without requiring any modification. Extensive experiments on multiple real-world datasets demonstrate that CoDCL consistently improves performance and exhibits strong generalizability across various temporal link prediction methods.
	\end{itemize}
	
	\section{Related Work}
	Dynamic graph representation learning has become a crucial research area and has been widely applied in many social fields \cite{9797305,yu2023towards}.
	These methods can extract complex dynamic behavioral patterns while maintaining temporal granularity, as detailed in references \cite{DBLP:conf/sigmod/WangLLXYWWCYSG21,tian2023freedyg,wang2025dynamic}. Various architectural innovations including memory-augmented networks \cite{11081903,DBLP:conf/sigir/0001GRTY20,10436118,DBLP:journals/corr/abs-2006-10637}, temporal random walk networks \cite{jin2022neural} and sequential learning networks \cite{DBLP:journals/corr/abs-2105-07944,cong2023do} have advanced the modeling of temporal interactions. Nevertheless, existing methodologies predominantly emphasize correlation-based learning models while neglecting fundamental causal relationships. Contrastive learning has been widely used to improve dynamic graph representations. DyTed \cite{zhang2023dyted} models dynamic networks via temporal-segment and structural contrastive learning. NeurTWs \cite{jin2022neural} designs interaction-based contrastive objectives to better capture spatiotemporal dynamics. TCL~\cite{DBLP:journals/corr/abs-2105-07944} applies contrastive learning by maximizing the mutual information between representations of future interacting nodes.
	
	Counterfactual-inspired augmentation aims to guide models to focus on causally relevant factors, thereby achieving stronger generalization ability \cite{9735356,11221992}. Existing research have introduced counterfactual reasoning into graph data scenarios, for example, GANITE uses counterfactual representation, and CFR enhances the recommendation effect through graph-level counterfactual augmentation. However, most of methods are limited to static graphs \cite{10564850,Huang_Liang_Huang_Zeng_Chen_Zhou_2025}. In contrast, for the complex and constantly changing temporal environment of dynamic graphs, related exploration is still in its infancy \cite{prenkaj2024unifying,qucody}, and existing works mostly focus on the interpretability of counterfactual reasoning. How to further apply counterfactual reasoning to the effective enhancement of temporal link prediction and representation learning still requires systematic research. 
	
	\section{Preliminary}\label{sec:preliminary}
	\noindent\textbf{Temporal Link Prediction:}
	Consider a continuous-time dynamic network $G = (\mathcal{V}, \mathcal{E}, \mathcal{T})$, where $\mathcal{V}$ represents the set of nodes, $\mathcal{E}$ denotes the set of edges, and $\mathcal{T}$ is the set of interaction times. Each edge $(u, v, t) \in \mathcal{E}$ represents the interaction relationship between nodes $u$ and $v$ at time $t$. We can obtain the node features $\mathbf{x}_u(t) \in \mathbb{R}^d$ at time $t$, where $d$ represents the feature dimension of the node attribute; and the edge features $\mathbf{e}_{uv}(t) \in \mathbb{R}^{d_e}$, where $d_e$ represents the feature dimension of the edge attribute.
	
	Historical interactions prior to a given deadline $t$ are denoted as $\mathcal{H}_t = \{(u,v,\tau) : \tau \leq t, (u,v,\tau) \in \mathcal{E}\}$. The temporal link prediction task aims to predict the probability that nodes $u$ and $v$ will form a link at a future time $t + \Delta t$:
	\begin{equation}
		\small
		P(y_{uv}(t + \Delta t) = 1 | \mathcal{H}_t),
	\end{equation}
	where $y_{uv}(t + \Delta t) \in \{0,1\}$ denotes the binary link indicator and $\Delta t$ represents the prediction time horizon.
	
	\noindent\textbf{Counterfactual data-enhanced Examples:}
	The counterfactual question is formulated as follows: For a given node pair $(u,v)$ with observed treatment $T_{uv}(t) = \tau$ and outcome $Y_{uv}(t) = y$, where $\tau$ denotes the observed treatment value and $y$ represents the observed outcome. Context $\mathbf{C}_{uv}(t)$ represents the learned representations of nodes $u$ and $v$, capturing their intrinsic attributes and historical interaction patterns. Treatment $T_{uv}(t)$ denotes a binary indicator representing the dynamic interaction pattern of nodes $u$ and $v$ at time $t$, defined through temporal structural properties. We ask: "What would the outcome $Y_{uv}(t)$ be if the treatment were $T_{uv}(t) = 1-\tau$ while keeping the context $\mathbf{C}_{uv}(t)$ unchanged?" This modeling encourages the model to learn causal relationships that are invariant to superficial temporal associations, leading to more robust and generalizable link prediction.
	\begin{figure*}[t]
		\centering
		\includegraphics[scale=0.45]{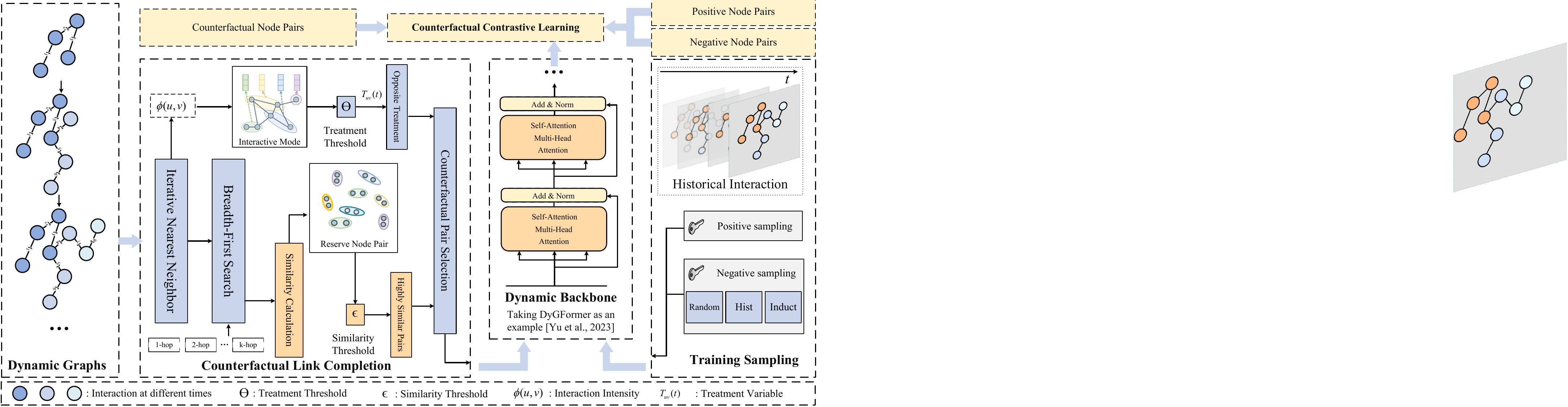}
		\caption{The framework of the proposed counterfactual-inspired augmentation dynamic network contrastive learning.}
		\label{4}
	\end{figure*}
	
	\section{Proposed Method}
	We propose a CoDCL framework that operates in two integrated stages: Counterfactual link Completion and Counterfactual Contrastive Learning, which jointly trains on factual and counterfactual examples to enhance temporal link prediction. Fig.2 illustrates the overall architecture of our approach. As shown in Algorithm 1, this is the pseudocode for the overall workflow of our proposed CoDCL framework, which primarily consists of Counterfactual Link Completion and Counterfactual Contrastive Learning.
	\subsection{Counterfactual Treatment Variables}
	\subsubsection{Dynamic Interaction Indicator}
	In dynamic network learning, we seek to improve the learning of node representations $\mathbf{z}_u(t)$ and $\mathbf{z}_v(t)$ by estimating the effect of treatment $T_{uv}(t)$ on outcome $y_{uv}(t+\Delta t)$ based on the counterfactual question ""Would the outcome be different if the treatment were different?"". And $\mathbf{z}_i(t)$ denotes the representation of node $i$ at time $t$ and $y_{uv}(t+\Delta t)$ represents the adjacency between nodes $u$ and $v$ at time $t+\Delta t$.
	
	Our counterfactual-inspired augmentation is grounded in defining meaningful treatment variables that capture temporal structural patterns. To this end, we propose a dynamic interaction indicator $T_{uv}(t)$ based on temporal common-neighbor analysis. To accommodate complex temporal variations, this measure quantifies the structural proximity between a node pair by measuring their shared neighborhood within a specified time window.
	
	Given a query time $t$ and time window parameter $\Delta$, we first define the time-constrained neighborhood for each node:
	\begin{equation}
		\small
		\mathcal{N}_u^{\Delta}(t) = \begin{cases} 
			\{w | \exists \tau \leq t, (u,w,\tau) \in \mathcal{E}\} & \text{if } \Delta = 0 \\
			\{w | \exists \tau \in [t-\Delta, t], (u,w,\tau) \in \mathcal{E}\} & \text{if } \Delta > 0,
		\end{cases}
	\end{equation}
	where we consider all historical interactions up to time $t$ when $\Delta = 0$. And when $\Delta > 0$, we focus on a sliding window $[t-\Delta, t]$ to capture recent interaction patterns.
	
	Subsequently, the interaction number $N(u,v,t)$ is used to measure the number of nodes that are common neighbors between nodes $u$ and $v$ within a specified time window:
	\begin{equation}
		\small
		N(u,v,t) = |\mathcal{N}_u^{\Delta}(t) \cap \mathcal{N}_v^{\Delta}(t)|.
	\end{equation}
	
	This indicator reflects the degree of structural overlap between node pairs, thus providing a fine-grained measure of the similarity of their interactions.
	
	To implement binary classification for counterfactual analysis, we set a global threshold based on interaction frequency patterns. For each node pair, we calculate the interaction intensity $\phi(u,v,t)$, which incorporates their historical interactions:
	\begin{equation}
		\small
		\phi(u,v,t) = \begin{cases}
			\sum_{(u,v,\tau) \in \mathcal{H}} w_{\tau} & \text{cumulative} \\
			\sum_{(u,v,\tau) \in \mathcal{H}} w_{\tau} \cdot e^{-\lambda(t_{\max} - \tau)} & \text{exponential decay},
		\end{cases}
	\end{equation}
	where $w_{\tau}$ denotes the weight associated with timestamp $\tau$, $\lambda$ represents the decay parameter, and $t_{\max}$ denotes the maximum timestamp in the dataset. The cumulative variant treats all interactions equally, while the exponential decay variant focuses more on recent interactions.

	The global treatment threshold is then determined using percentile-based statistics methods:
	\begin{equation}
		\small
		\theta = \text{Percentile}(\{\phi(u,v,t) | \forall (u,v) \in \mathcal{E}\}, p),
	\end{equation}
	where $p$ represents the percentile parameter. This global threshold adapts to the specific characteristics of different datasets, ensuring that the distinction captures meaningful interaction patterns rather than random variations.
	
	\subsubsection{Binary Treatment}
	Obtaining the binary treatment requires comparing the interaction intensity with a global threshold:
	\begin{equation}
		\small
		T_{uv}(t) = \mathbb{I}[\phi(u,v,t) \geq \theta],
	\end{equation}
	where $\mathbb{I}[\cdot]$ denotes the indicator function. This binary treatment distinguishes node pairs with high interaction frequency from those with low interaction frequency, thus realizing the counterfactual operation set in the algorithm.
	
	The synergy between the continuous interaction indicator $T_{uv}(t)$ and the binary treatment classification provides a comprehensive framework for counterfactual-inspired augmentation. 
	The continuous indicator captures fine-grained structural relationships, while the binary classification, by comparing node pairs with different treatments but similar structural features, plays a crucial role in enhancing counterfactual data in dynamic networks.
	
	\subsection{Counterfactual Link Completion Strategy}
	
	\subsubsection{Nearest neighbor breadth search}
	To effectively identify counterfactual pairs within local neighborhood structures, we employ a nearest-neighbor breadth search strategy, which systematically explores the $k$-hop neighborhood of the target node. 
	For each factual node pair $(u,v)$ with $T_{uv}(t)$, we begin the search from their direct neighbors and progressively expand the search radius.
	
	This search process is performed in iterative layers. First, we examine the 1-hop neighborhoods $\mathcal{N}_u^1(t)$ and $\mathcal{N}_v^1(t)$ to construct candidate pairs. Within each k-hop layer, we generate candidate pairs to achieve maximum coverage $(u', v')$, where $u' \in \mathcal{N}_u^k(t)$ and $v' \in \mathcal{N}_v^k(t)$. We use binary classification $T_{u'v'}(t) \neq T_{uv}(t)$ and similarity constraints to evaluate the treatment opposition for each candidate pair.
	
	This breadth-first expansion ensures that we prioritize structurally proximate candidates before exploring distant nodes, thus maintaining computational efficiency while guaranteeing the semantic relevance of the identified counterfactual pairs. The search terminates either when a valid counterfactual pair satisfying all constraints is found, or when the maximum hop count limit $k_{\max}$ is reached.
	
	\subsubsection{Counterfactual Pair Selection}
	To find suitable counterfactual pairs, we need representations that can capture both the intrinsic properties of nodes and their structural roles. For each node $u$ at time $t$, we perform the following calculation:
	\begin{equation}
		\small
		h_u(t) = f_{s}(\mathbf{x}_u(t)) + \frac{1}{|\mathcal{N}_u(t)|} \sum_{w \in \mathcal{N}_u(t)} f_{n}(\mathbf{x}_w(t)),
	\end{equation}
	where $f_{s}$ and $f_{n}$ denote transformation functions, and $\mathcal{N}_u(t)$ represents the temporal neighbors of node $u$ before time $t$. 
	This modeling approach combines the features of a single node with the influence of its neighborhood, ensuring that similarity matching considers both intrinsic attributes and structural context, while maintaining computational efficiency.
	
	For each $(u,v)$ with interaction indicator $T_{uv}(t)$, we identify its counterfactual pair $(\hat{u}, \hat{v})$ through the optimization:
	\begin{equation}
		\small
		(\hat{u}, \hat{v}) = \arg\max_{(u',v') \in \mathcal{C}_k(t)} \frac{1}{2}[\cos(h_u(t), h_{u'}(t)) + \cos(h_v(t), h_{v'}(t))],
	\end{equation}
	where opposite treatment $T_{u'v'}(t) \neq T_{uv}(t)$ and k-hop structural proximity $(u',v') \in \mathcal{N}_u^k(t) \times \mathcal{N}_v^k(t)$ for some $k \leq k_{\max}$; and candidate validity $(u',v') \in \mathcal{C}_k(t)$, $\mathcal{C}_k(t)$ represents the set of candidate node pairs within $k$-hop neighborhoods with opposite treatment classification at time $t$. The cosine similarity function $\cos(\cdot, \cdot)$ ensures counterfactual pairs are close in the representation space.
	
	\begin{algorithm}[h]
		\clearpage
		\small
		\caption{Counterfactual-inspired augmentation Dynamic network Contrastive Learning}
		\label{alg:counterfactual_complete_learning}
		\begin{algorithmic}[1]
			\Require Temporal graph $\mathcal{G}$, node features $\mathbf{X}$, time window $\Delta$, percentile $p$, max hop $k_{\max}$, decay parameter $\lambda$
			\Require Loss balance $\alpha$, temperature $\tau$, learning rate $\eta$
			\Ensure Trained model $\Theta = \{\text{DynamicBackbone}, \text{LinkPredictor}\}$
			
			\For{each node pair $(u,v) \in \mathcal{E}$}
			\State $\phi(u,v) \leftarrow \sum_{(u,v,\tau) \in \mathcal{H}} w_{\tau} \cdot e^{-\lambda(t_{\max} - \tau)}$
			\EndFor
			\State $\theta \leftarrow \text{Percentile}(\{\phi(u,v) | \forall (u,v) \in \mathcal{E}\}, p)$	
			\State // \textbf{Generate Treatment Variables}
			\For{each query time $t$ and node pair $(u,v)$}
			\If{$\Delta = 0$}
			\State $\mathcal{N}_u^{\Delta}(t) \leftarrow \{w | \exists \tau \leq t, (u,w,\tau) \in \mathcal{E}\}$
			\Else
			\State $\mathcal{N}_u^{\Delta}(t) \leftarrow \{w | \exists \tau \in [t-\Delta, t], (u,w,\tau) \in \mathcal{E}\}$
			\EndIf
			\State $T_{uv}(t) \leftarrow \mathbb{I}[\phi(u,v,t) \geq \theta]$
			\EndFor
			\State // \textbf{Nearest Neighbor Breadth Search for Counterfactuals}
			\For{each node $u$ at time $t$}
			\State $h_u(t) \leftarrow f_s(\mathbf{x}_u(t)) + \frac{1}{|\mathcal{N}_u(t)|} \sum_{w \in \mathcal{N}_u(t)} f_n(\mathbf{x}_w(t))$
			\EndFor
			\For{each factual pair $(u,v,t)$ with treatment $T_{uv}(t)$}
			\For{$k = 1$ to $k_{\max}$}
			\State // Generate candidate pairs
			\State $\mathcal{C}_k \leftarrow \{(u',v') | u' \in \mathcal{N}_u^k, v' \in \mathcal{N}_v^k, T_{u'v'}(t) \neq T_{uv}(t)\}$
			\If{$\mathcal{C}_k\neq \emptyset$}
			
			\State $(\hat{u}, \hat{v}) \leftarrow \arg\max_{(u',v') \in \mathcal{C}_k} \frac{1}{2} \cdot [\cos(h_u, h_{u'}) + \cos(h_v, h_{v'})]$
			\State \textbf{break}
			\EndIf
			\EndFor
			\State Store mapping: $(u,v) \rightarrow (\hat{u}, \hat{v})$
			\EndFor
			
			\State // \textbf{Counterfactual Contrastive Learning}
			\State Initialize $\Theta$ and AdamW optimizer
			
			\For{each training epoch}
			\For{each batch $\mathcal{B} = \{(u_i, v_i, t_i, \hat{u}_i, \hat{v}_i)\}$}
			
			\State // \textbf{Compute Edge Representations}
			\For{$i = 1$ to $|\mathcal{B}|$}
			\State $h_{u_i}, h_{v_i} \leftarrow \text{DynamicBackbone}(\mathcal{G}, u_i, v_i, t_i)$
			\State $z^{\text{pos}}_i,z^{\text{cf}}_i,z^{\text{neg}}_i \leftarrow h_{u_i} \cdot h_{v_i},h_{\hat{u}_i} \cdot h_{\hat{v}_i},h_{u'_i} \cdot h_{v'_i}$
			\EndFor
			
			\State // \textbf{Compute Losses}
			\State $\mathcal{L}_{\text{f}} \leftarrow -\sum_i [\log \sigma(s^{\text{pos}}_i) + \log(1-\sigma(s^{\text{neg}}_i))]$
			\State $\mathcal{L}_{\text{c}} \leftarrow -\sum_i \log \frac{\exp(\cos(\psi, z^{\text{cf}}_i)/\tau)}{\exp(\cos(z^{\text{pos}}_i, z^{\text{cf}}_i)/\tau) + \exp(\cos(z^{\text{pos}}_i, z^{\text{neg}}_i)/\tau)}$
			\State $\mathcal{L}_{\text{total}} \leftarrow \alpha \cdot \mathcal{L}_{\text{fact}} + (1-\alpha) \cdot \mathcal{L}_{\text{contrast}}$
			
			\State // \textbf{Update Parameters}
			\State $\Theta \leftarrow \Theta - \eta \nabla_{\Theta} \mathcal{L}_{\text{total}}$ with gradient clipping
			\EndFor
			\EndFor
		\end{algorithmic}
	\end{algorithm}
	\subsection{Counterfactual Contrastive Learning}
	\subsubsection{Model Architecture}
	Our counterfactual contrastive learning module includes a dynamic backbone and the final output. The dynamic backbone is broadly applicable and plug-and-play, seamlessly integrating into existing temporal architectures such as TGAT, GraphMixer, and DyGFormer without modification. The backbone outputs the node embedding $h_u(t)$ of each node $u$ at time $t$, which is then used for the final mapping, transforming edge representations into link probabilities. This is implemented as a multilayer perceptron with residual connections and batch normalization to ensure the stability of the training dynamics.
	
	For each training batch $\mathcal{B} = \{(u_i, v_i, t_i, \hat{u}_i, \hat{v}_i)\}_{i=1}^{|\mathcal{B}|}$, where $|\mathcal{B}|$ denotes the batch size, we compute three types of edge representations through our multi-level embedding computation. The positive edge representation for truly existing links:
	\begin{equation}
		\small
		z^{\text{pos}}_i = h_{u_i}(t_i) \cdot h_{v_i}(t_i).
	\end{equation}
	\begin{equation}
		\small
		z^{\text{cf}}_i = h_{\hat{u}_i}(t_i) \cdot h_{\hat{v}_i}(t_i).
	\end{equation}
	\begin{equation}
		\small
		z^{\text{neg}}_i = h_{u'_i}(t_i) \cdot h_{v'_i}(t_i),
	\end{equation}
	where $(u'_i, v'_i)$ represents negative samples drawn using standard sampling strategies, and $(\hat{u}_i, \hat{v}_i)$ denotes the counterfactual pair identified through our selection procedure. 
	
	\subsubsection{Loss Function}
	Our training objective combines factual and counterfactual prediction through a contrastive learning framework. The factual link prediction loss is defined as:
	\begin{equation}
		\small
		\mathcal{L}_{\text{f}}
		=
		-\sum_{i=1}^{|\mathcal{B}|}
		\left[
		\log \sigma(s_i^{\text{pos}})
		+
		\log \left(1-\sigma(s_i^{\text{neg}})\right)
		\right],
	\end{equation}
	where $s^{\text{pos}}_i$, $s^{\text{neg}}_i$ denote prediction scores of $z^{\text{pos}}_i$ and $z^{\text{neg}}_i$.
	The counterfactual contrastive loss employs an InfoNCE-based framework:
	\begin{equation}
		\small
		\mathcal{L}_{\text{c}} =\sum_{i=1}^{|\mathcal{B}|} \log \frac{\exp(\cos(\psi, z^{\text{cf}}_i)/\tau)}{\exp(\cos(z^{\text{pos}}_i, z^{\text{cf}}_i)/\tau) + \exp(\cos(z^{\text{pos}}_i, z^{\text{neg}}_i)/\tau)},
	\end{equation}
	where $\psi$ is determined by the actual state of the counterfactual pair. If the actual state exists, $\psi$ is $z^{\text{pos}}_i$, otherwise it is $z^{\text{neg}}_i$, and the direction of comparison is determined by the actual state. $\tau > 0$ represents the temperature parameter controlling the distribution concentration, and the cosine similarity function is defined as $\cos(z_1, z_2) = \frac{z_1 \cdot z_2}{\|z_1\|_2 \cdot \|z_2\|_2}$.
	The joint optimization objective balances factual prediction and counterfactual reasoning:
	\begin{equation}
		\small
		\mathcal{L}_{\text{total}} = \alpha \cdot \mathcal{L}_{\text{f}} + (1-\alpha) \cdot \mathcal{L}_{\text{c}},
	\end{equation}
	where $\alpha \in [0,1]$ balances factual prediction and counterfactual reasoning.
	
	\subsubsection{Algorithm Complexity}	
	Counterfactual Link Completion involves computing common-neighbor statistics and performing a bounded $k$-hop neighborhood search to generate counterfactual pairs, with an overall complexity of $\mathcal{O}(|E|\, d_{\text{avg}}^{k})$ when the number of processed node pairs is on the same order as the number of observed edges. Here, $|E|$ is the number of edges, $d_{\text{avg}}$ is the average node degree, and $k$ is the maximum search depth. Specifically, computing $T_{ij}$ via common-neighbor set intersection costs $\mathcal{O}(d_{\text{avg}})$ on average for each pair, while the bounded $k$-hop search costs $\mathcal{O}(d_{\text{avg}}^{k})$ in the worst case and dominates the overall complexity. Counterfactual contrastive training computes temporal embeddings for positive, negative, and counterfactual samples, followed by contrastive learning. Compared with standard dynamic backbone training, this stage introduces only a small constant-factor overhead. Therefore, the main computational cost of the proposed framework is concentrated in the counterfactual pair generation phase, which can be performed offline in advance to reduce the burden of the subsequent training process.
	\section{Experimental Design and Evaluation}
	\subsection{Experimental Data and Comparative Methods}
	This study employs seven real-world datasets for experimental evaluation, including community interaction networks such as Wikipedia and Reddit, online education data from MOOC, user behavior data from LastFM, email communication network data from Enron, academic network data from UCI, and political interaction network data from CanParl~\cite{poursafaei2022towards}. Among them, LastFM and Reddit are relatively larger in scale. Detailed statistics of these real-world datasets are provided in Table \ref{tab:data_statistics}.
	
	\textbf{Wikipedia:} represents a temporal bipartite network capturing user editing activities on Wikipedia pages. The network structure comprises user nodes and page nodes, with temporal edges representing editing events.
	
	\textbf{Reddit:} constitutes a temporal bipartite graph modeling user posting behaviors across various subreddits during a monthly observation period.
	
	\textbf{MOOC:} forms a bipartite temporal network documenting student engagement patterns with educational content in online learning environments. Students and instructional materials serve as network entities.
	
	\textbf{LastFM:} constructs a bipartite temporal graph representing music consumption patterns between users and songs. 
	
	\textbf{Enron:} establishes a temporal communication network derived from email exchanges among employees within the ENRON corporation spanning a three-year operational period.
	
	\textbf{UCI:} models an academic communication network where university students form nodes and their message exchanges constitute temporal edges.
	
	\textbf{CanParl:} encodes a political interaction network tracking legislative collaboration among Canadian Members of Parliament. Parliamentary representatives from electoral constituencies form network nodes.
	
	\begin{table}[!htbp]
		\centering
		\caption{Statistics of the datasets.}
		\label{tab:data_statistics}
		\begin{tabular}{c|cccc}
			\hline
			Datasets & \#Nodes & \#Links & Duration & Unique Steps \\ \hline
			UCI & 1,899 & 59,835 & 196 days & 58,911\\
			CanParl & 734 & 74,478 & 14 years & 14  \\ 
			Enron & 184 & 125,235 & 3 years & 22,632  \\
			Wikipedia & 9,227 & 157,474 & 1 month & 152,757  \\
			MOOC & 7,144 & 411,749 & 17 months & 345,600  \\
			Reddit & 10,984 & 672,447 & 1 month & 669,065  \\
			LastFM & 1,980 & 1,293,103 & 1 month & 1,283,614 \\
			\hline
		\end{tabular}
	\end{table}
	
	Nine state-of-the-art methods are systematically compared as benchmarks, covering a variety of technical approaches. These benchmarks include JODIE~\cite{DBLP:conf/kdd/KumarZL19}, DyRep~\cite{DBLP:conf/iclr/TrivediFBZ19}, CAWN~\cite{DBLP:conf/iclr/WangCLL021}, TGN~\cite{DBLP:journals/corr/abs-2006-10637}, TCL~\cite{DBLP:journals/corr/abs-2105-07944}, GraphMixer~\cite{cong2023do}, DyGFormer~\cite{yu2023towards}, FreeDyG~\cite{tian2023freedyg}, and CorDGT~\cite{wang2025dynamic}, all of which represent the current state-of-the-art in the field of continuous-time dynamic link prediction.
	
	The evaluation uses Average Precision (AP) and Area Under the Receiver Operating Characteristic Curve (AUC-ROC) metrics. All datasets are chronologically divided into training, validation, and test sets with proportions of 70\%, 15\%, and 15\% respectively. And all experiments are conducted on a machine equipped with Intel Xeon Gold 6326 CPU at 2.90GHz and NVIDIA RTX A6000.
	
	\subsection{Experimental Complexity Analysis}
	To test the time complexity of the model we proposed, we conducted an experimental complexity analysis. Mechanistically, CoDCL consists of two stages: \textbf{counterfactual node-pair generation} and \textbf{model training}. We have already analyzed the complexity of the first stage in IV.C 3), which constitutes the main computational cost of the model. However, this stage can be completed in advance during pre-training or offline preprocessing, thereby reducing the computational burden in subsequent training processes. Meanwhile, the first stage is an offline preprocessing step before training. The sensitivity study in Section V.F shows that performance is already saturated at $k_{\max}=2$, suggesting that large search depths are unnecessary in practice and thus limiting the actual BFS overhead. After obtaining the counterfactual pairs, the training stage proceeds with standard backpropagation. To clarify the efficiency, we further provide a bar chart of training time, reporting the per-epoch model training before and after adding CoDCL. As shown in Figure, once the counterfactual pairs are constructed, CoDCL introduces only one additional contrastive branch over the original backbone, resulting in relatively limited extra model training cost. The training phase complexity differs little from standard dynamic backbone training.
	\begin{figure}[h]
		\centering
		\includegraphics[width=0.49\linewidth]{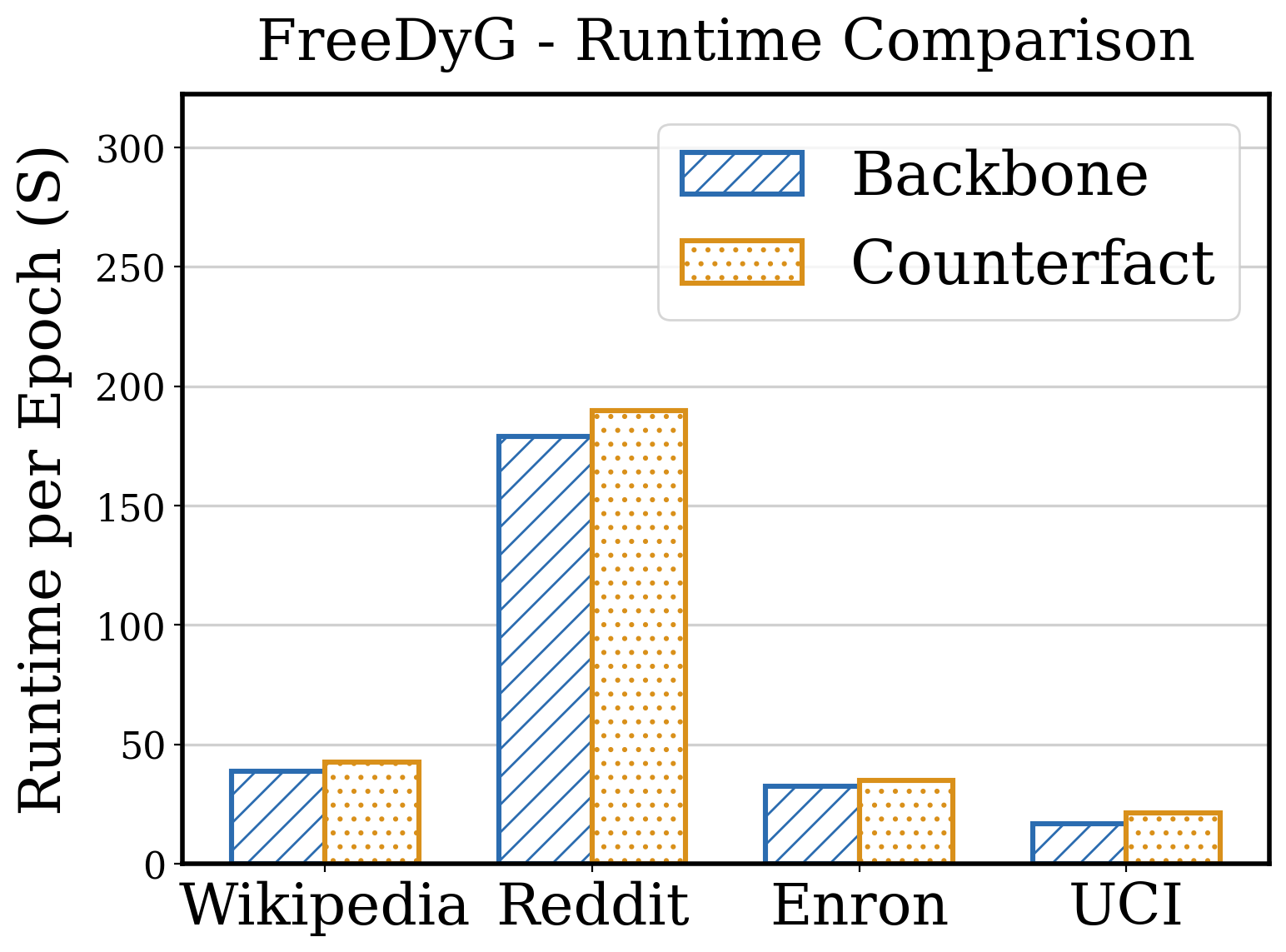}
		\includegraphics[width=0.49\linewidth]{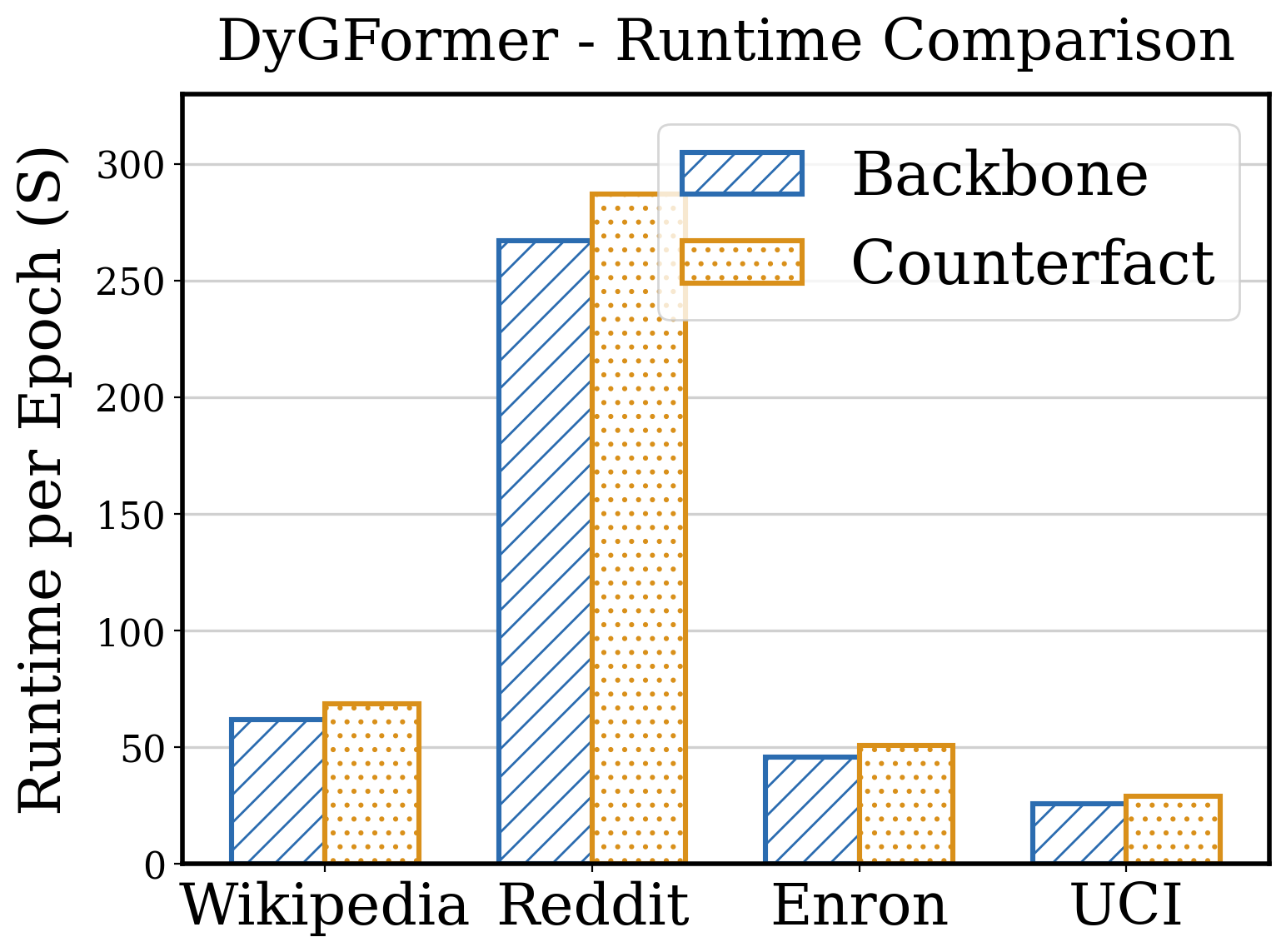}
		\caption{Comparison of the running time per epoch between the counterfactual and the backbone on different datasets.}
		\label{1}
	\end{figure}
	
	\subsection{Experimental Results Analysis}
	\begin{table*}[h]
		\centering
		\caption{Performance comparison under transductive settings with AP using three backbone architectures for CoDCL evaluation.}
		\label{tab:performance_comparison}
		\small
		\resizebox{\textwidth}{!}{
			\begin{tabular}{c|cccccccc}
				\toprule
				\midrule
				&\textbf{Method} & \textbf{Wikipedia} & \textbf{UCI} & \textbf{Enron} & \textbf{MOOC} & \textbf{Reddit} & \textbf{LastFM} & \textbf{CanParl} \\
				\midrule
				\multirow{5}{*}{Baselines}
				&JODIE & 96.50±0.14 & 89.43±1.09 & 84.77±0.30 & 80.23±2.44 & 98.31±0.14 & 70.85±2.13 & 69.26±0.31 \\
				&DyRep &  94.86±0.06   &  65.14±2.30   &   82.38±3.36  &  81.97±0.49   &   98.22±0.04  & 71.92±2.21    &  66.54±2.76   \\
				&TCL &   96.47±0.16  &  89.57±1.63   &  79.70±0.71   &  82.38±0.24   &  97.53±0.02   & 67.27±2.16    & 68.67±2.67    \\     
				&TGN & 98.45±0.06 & 92.34±1.04 & 86.53±1.11 & {89.15±1.60} & 98.63±0.06 & 77.07±3.97 & 70.88±2.34 \\
				&CAWN & 98.76±0.03 & 95.18±0.06 & 89.56±0.09 & 80.15±0.25 & 99.11±0.01 & 86.99±0.06 & 69.82±2.34 \\
				&GraphMixer & 97.17±0.05 & 93.50±0.49 & 81.08±0.73 & 82.73±0.16 & 97.37±0.01 & 75.64±0.23 & 74.60±0.37 \\
				\midrule
				\multirow{3}{*}{{DyGFormer}}
				&Backbone & 98.75±0.02 & 95.76±0.15 & 92.43±0.12 & 87.23±0.45 & 99.11±0.02 & 92.07±0.28 & 97.36±0.45 \\
				&CoDCL & \textbf{99.13±0.01} & \textbf{96.28±0.08} & \textbf{92.83±0.08} & \textbf{87.84±0.33} & \textbf{99.35±0.02} & \textbf{93.15±0.16} & \textbf{98.46±0.41} \\
				\midrule
				\multirow{3}{*}{{FreeDyG}}
				&Backbone & 99.23±0.03 & 96.03±0.12 & 92.51±0.05 & 88.82±0.43 & 99.24±0.02 & 90.74±0.09 & 74.52±0.71 \\
				&CoDCL & \textbf{99.72±0.01} & \textbf{96.73±0.09} & \textbf{92.89±0.21} & \textbf{89.17±0.40} & \textbf{99.63±0.01} & \textbf{90.85±0.05} & \textbf{76.73±0.64} \\
				\midrule
				\multirow{3}{*}{{CorDGT}}
				&Backbone & 99.05±0.01 & 96.03±0.02 & 91.76±0.50 & 84.35±0.44 & 99.02±0.01 & 92.23±0.10 & 69.87±0.62 \\
				&CoDCL & \textbf{99.24±0.02} & \textbf{96.94±0.02} & \textbf{92.31±0.26} & \textbf{85.12±0.55} & \textbf{99.42±0.03} & \textbf{92.78±0.32} & \textbf{71.44±0.68} \\
				\midrule
				\bottomrule
		\end{tabular}}
	\end{table*}
	\begin{table*}[h]
		\centering
		\caption{Performance comparison under inductive settings with AP using three backbone architectures for CoDCL evaluation.}
		\label{tab:inductive_link_prediction}
		\small
		\resizebox{\textwidth}{!}{%
			\begin{tabular}{c|cccccccc}
				\toprule
				\midrule
				&\textbf{Method} & \textbf{Wikipedia} & \textbf{UCI} & \textbf{Enron} & \textbf{MOOC} & \textbf{Reddit} & \textbf{LastFM} & \textbf{CanParl} \\
				\midrule
				\multirow{5}{*}{Baselines}
				&JODIE & 94.82±0.20 & 79.86±1.48 & 80.72±1.39 & 79.63±1.92 & 96.50±0.13 & 81.61±3.82 & 53.92±0.94 \\
				&DyRep & 92.43±0.37& 57.48±1.87    & 74.55±3.95    &   81.07±0.44  &  96.09±0.11   & 83.02±1.48    & 54.02±0.76    \\
				&TCL & 96.22±0.17    &  87.36±2.03   &   76.14±0.79  & 80.60±0.22    & 94.09±0.07    & 73.53±1.66    & 54.30±0.66    \\ 
				&TGN & 97.83±0.04 & 88.12±2.05 & 77.94±1.02 & {89.04±1.17} & 97.50±0.07 & 81.45±4.29 & 54.10±0.93 \\
				&CAWN & 98.24±0.03 & {92.73±0.06} & 86.35±0.51 & 81.42±0.24 & 98.62±0.01 & 89.42±0.07 & 55.80±0.69 \\
				&GraphMixer & 96.65±0.02 & 91.19±0.42 & 75.88±0.48 & 81.41±0.21 & 95.26±0.02 & 82.11±0.42 & 55.91±0.82 \\
				\midrule
				\multirow{3}{*}{{DyGFormer}}
				&Backbone & 98.59±0.03 & {94.45±0.12} & 89.76±0.34 & 86.96±0.43 & {98.63±0.02} & {94.09±0.11} & 87.74±0.71 \\
				&CoDCL & \textbf{98.86±0.01} & \textbf{95.24±0.08} & \textbf{90.54±0.20} & \textbf{87.29±0.40} & \textbf{98.97±0.01} & \textbf{94.33±0.11} & \textbf{90.85±0.64} \\
				\midrule
				\multirow{3}{*}{{FreeDyG}}
				&Backbone & 98.91±0.03 & {94.50±0.12} & 89.69±0.17 & 88.05±0.43 & 98.84±0.02 & 93.01±0.09 & 55.28±1.24 \\
				&CoDCL & \textbf{99.14±0.01} & \textbf{94.78±0.43} & \textbf{90.17±0.48} & \textbf{88.72±0.11} & \textbf{99.35±0.01} & \textbf{93.82±0.29} & \textbf{58.17±1.01} \\
				\midrule
				\multirow{3}{*}{{CorDGT}}
				&Backbone & 98.48±0.02 & 94.70±0.10 & 91.65±0.13 & 83.74±0.51 & 97.82±0.02 & 93.03±0.18 & 53.22±0.76  \\
				&CoDCL & \textbf{98.96±0.02} & \textbf{95.48±0.52} & \textbf{92.88±0.12} & \textbf{84.59±0.27} & \textbf{98.33±0.02} & \textbf{93.91±0.22} & \textbf{55.46±0.74} \\
				\midrule
				\bottomrule
		\end{tabular}}
	\end{table*}
	
	Table \ref{tab:performance_comparison} and Table \ref{tab:inductive_link_prediction} evaluates the CoDCL framework on seven datasets using three different backbone architectures. Results, reported as mean $\pm$ standard deviation over multiple runs.
	Our method consistently improves performance when integrated with DyGFormer, FreeDyG, and CorDGT. Specifically, it improves performance by 0.4\% on Reddit and by 1.57\% on CanParl. This indicates that counterfactual data augmentation captures fundamental temporal interaction patterns rather than network architecture-specific phenomena. Table \ref{tab:inductive_link_prediction} reports inductive evaluation results on unseen nodes. The framework remains robust across various network types. Performance improvements vary by domain, with larger improvements on communication networks and smaller but equally stable improvements on community interaction networks, suggesting that the method is particularly effective under complex and irregular temporal dependencies.
	
	\begin{figure}[h]
		\centering
		\begin{subfigure}{0.23\textwidth}
			\includegraphics[width=\textwidth]{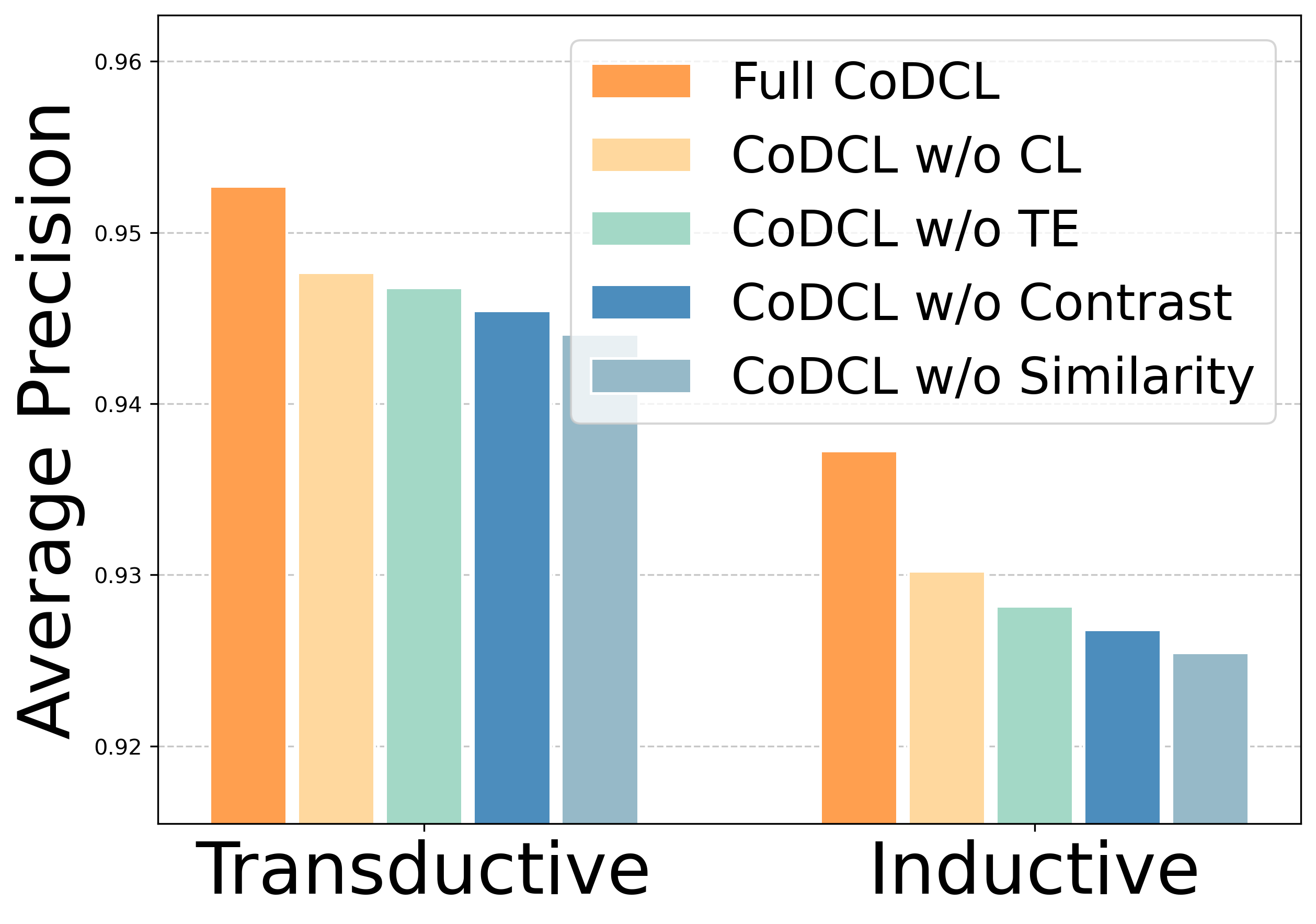}
		\end{subfigure}
		\begin{subfigure}{0.23\textwidth}
			\includegraphics[width=\textwidth]{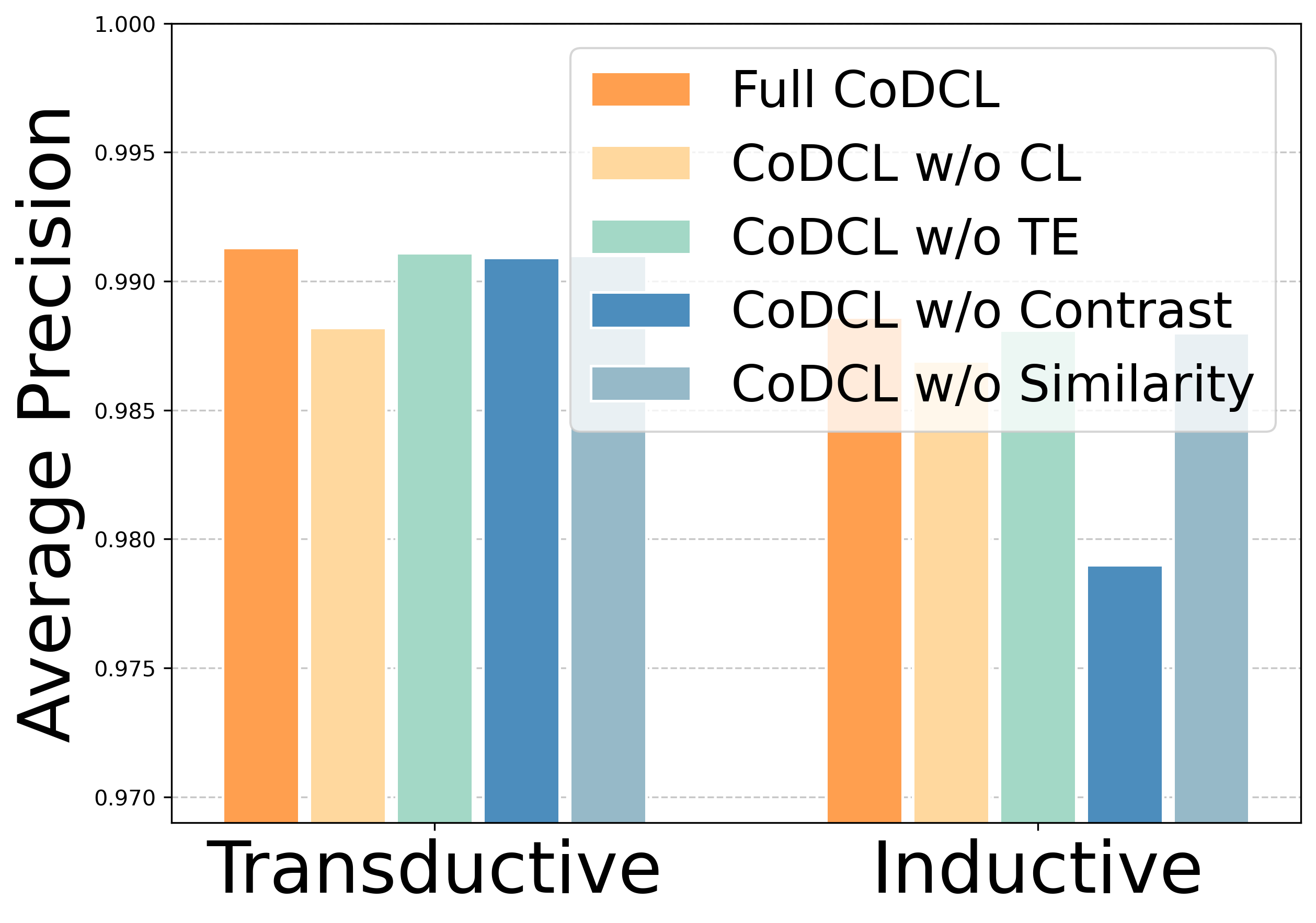}
		\end{subfigure}
		\begin{subfigure}{0.23\textwidth}
			\includegraphics[width=\textwidth]{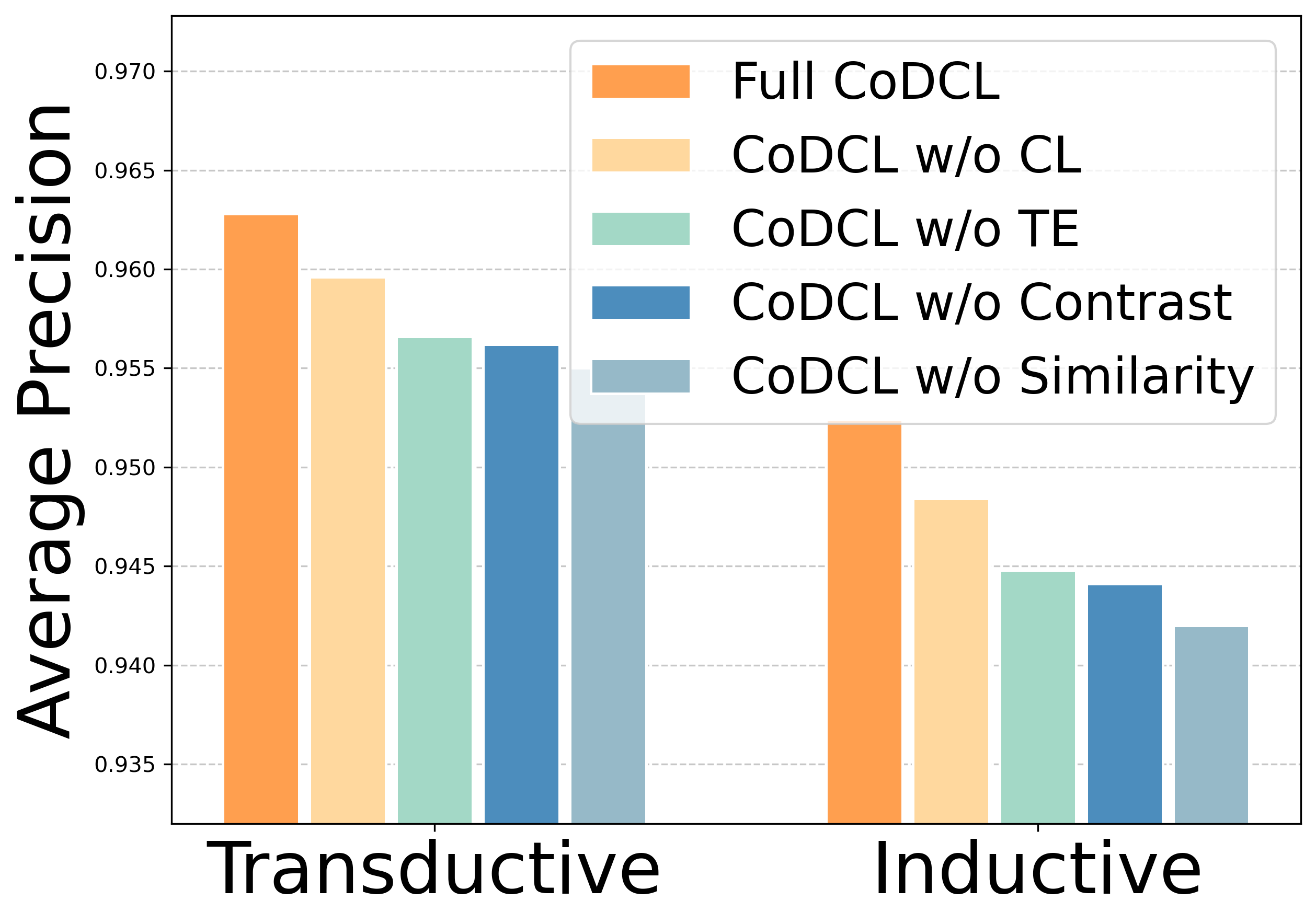}
		\end{subfigure}
		\begin{subfigure}{0.23\textwidth}
			\includegraphics[width=\textwidth]{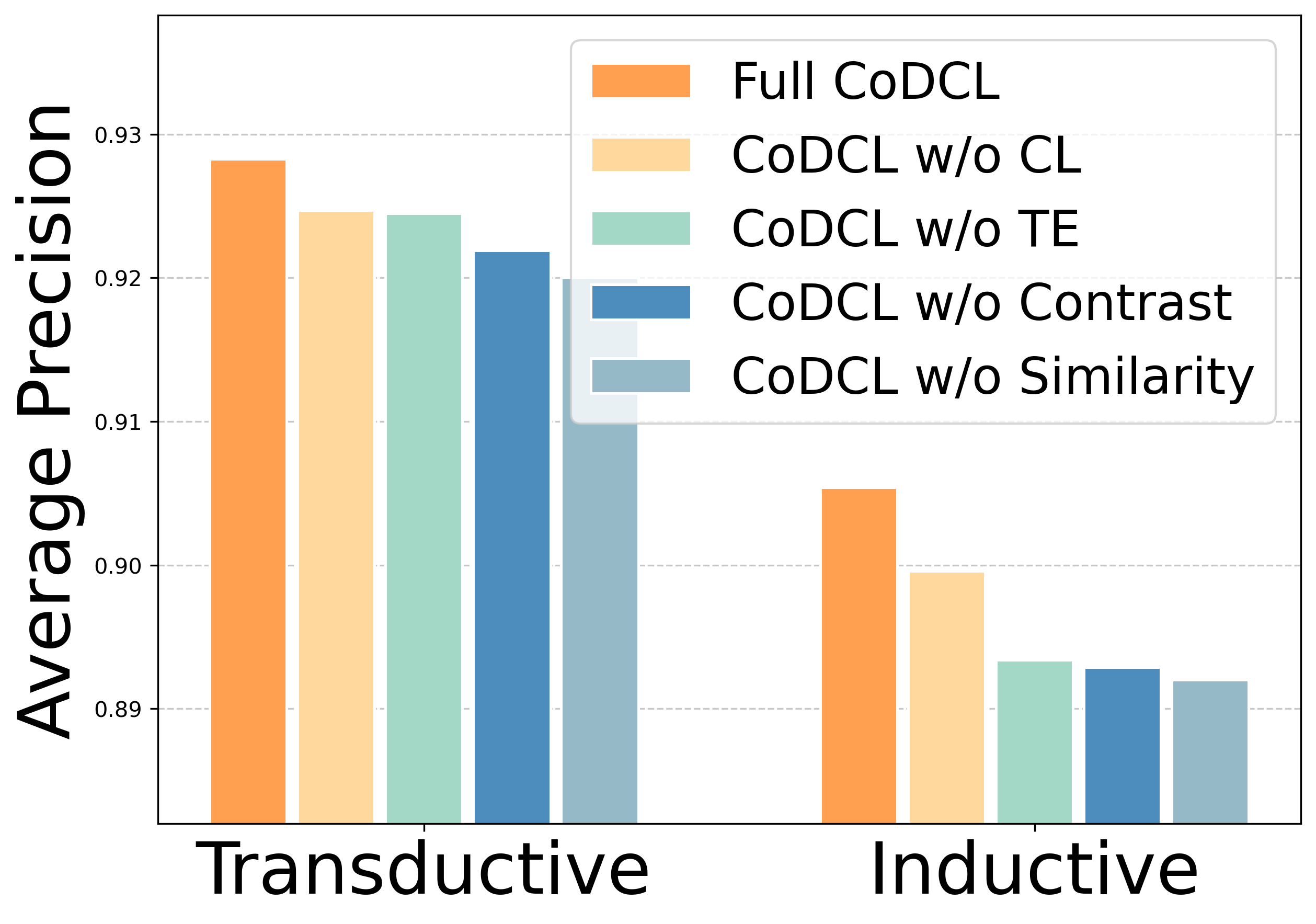}
		\end{subfigure}
		\begin{subfigure}{0.23\textwidth}
			\includegraphics[width=\textwidth]{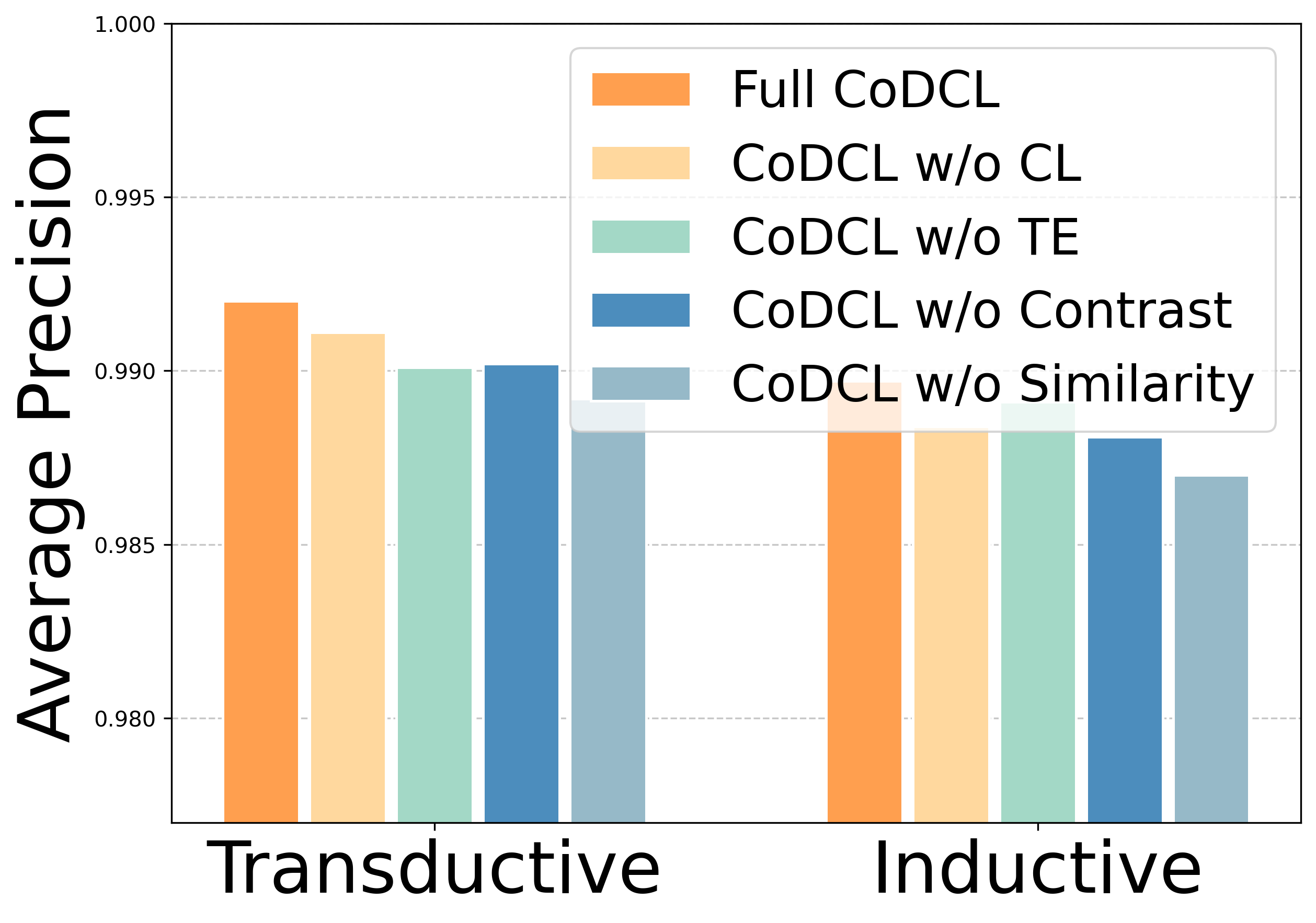}
		\end{subfigure}
		\begin{subfigure}{0.23\textwidth}
			\includegraphics[width=\textwidth]{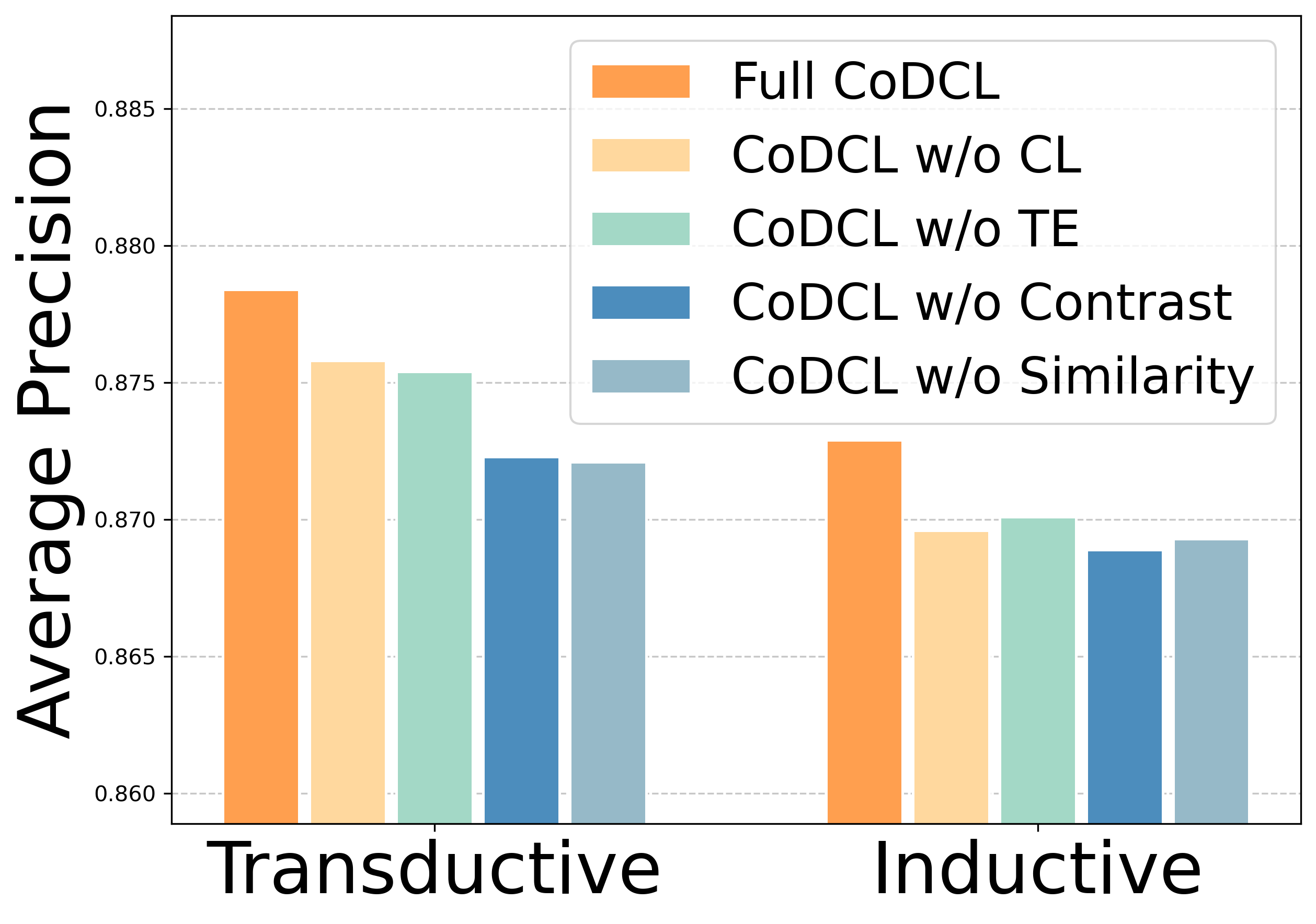}
		\end{subfigure}
		\begin{subfigure}{0.23\textwidth}
			\includegraphics[width=\textwidth]{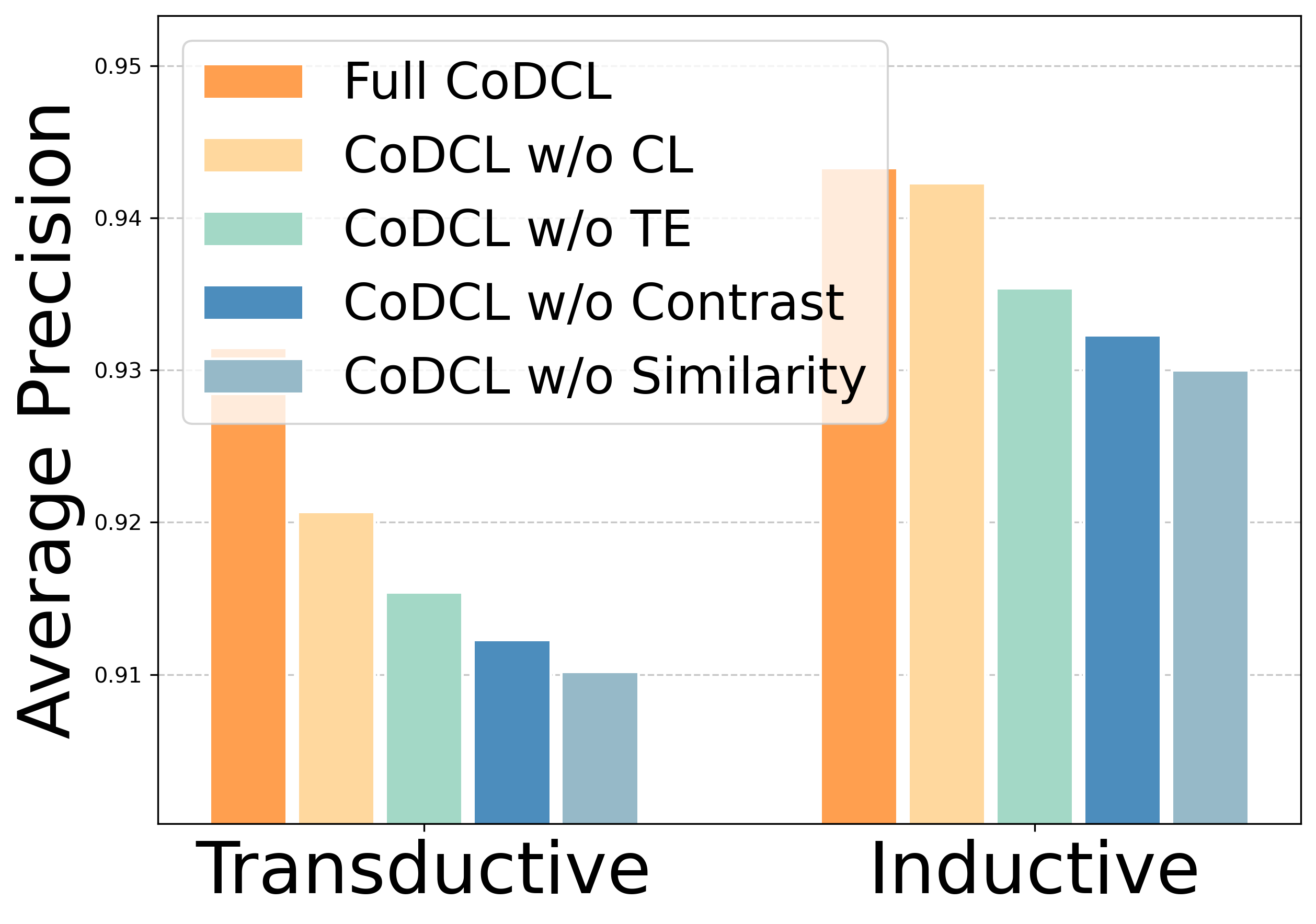}
		\end{subfigure}
		\begin{subfigure}{0.23\textwidth}
			\includegraphics[width=\textwidth]{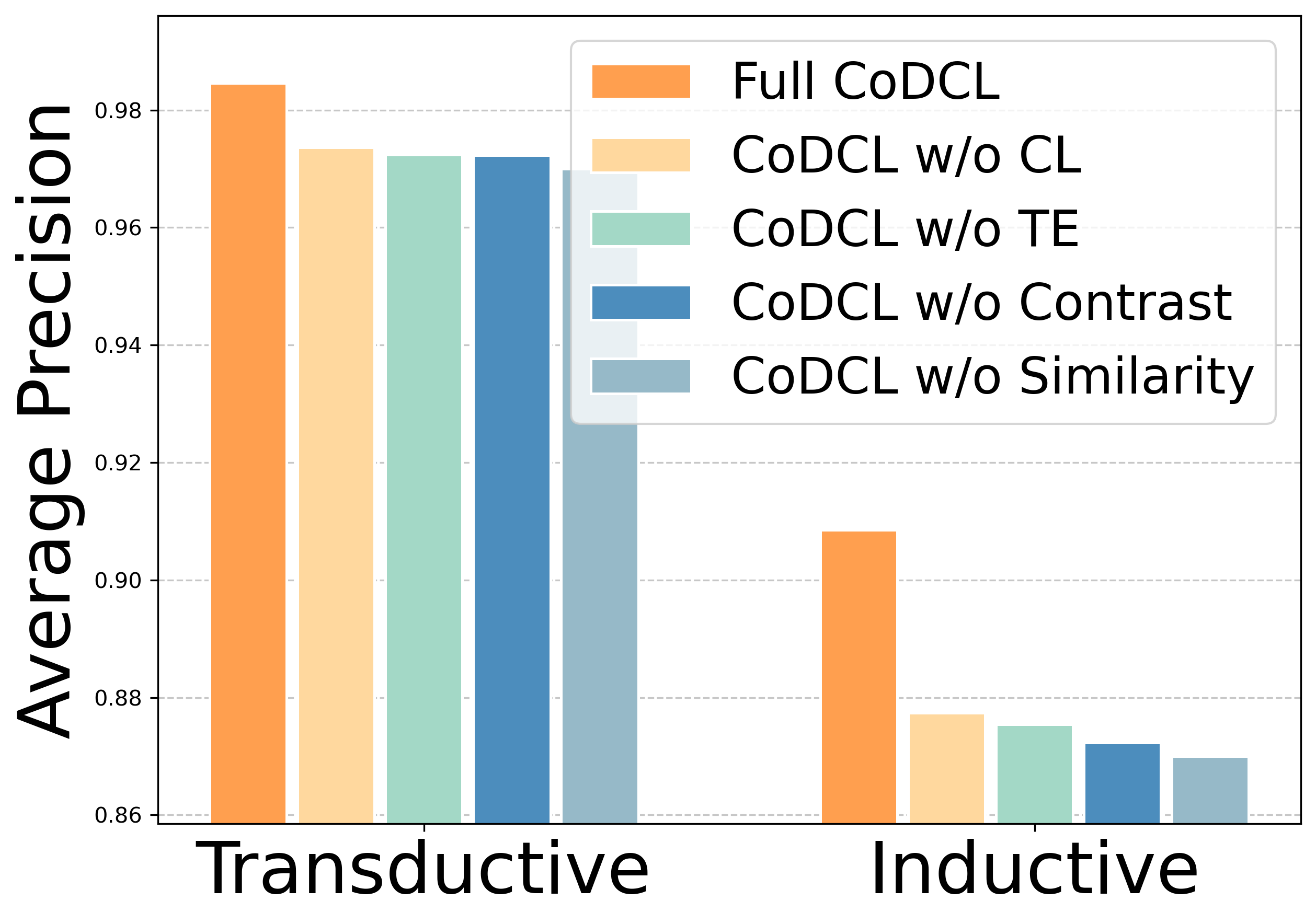}
		\end{subfigure}
		\caption{Ablation study and their average results under both transductive and inductive settings by removing: Counterfactual Learning (w/o CL), Temporal Encoding (w/o TE), Contrastive learning (w/o Contrast), and similarity constraints (w/o Similarity). (a) Average, (b) Wikipedia, (c) UCI, (d) Enron, (e) Reddit, (f) Mooc, (g) LastFM, (h) CanParl.}
		\label{fig:ablation}
	\end{figure}
	\begin{figure}[t]
		\centering
		\begin{subfigure}{0.23\textwidth}
			\includegraphics[width=\textwidth]{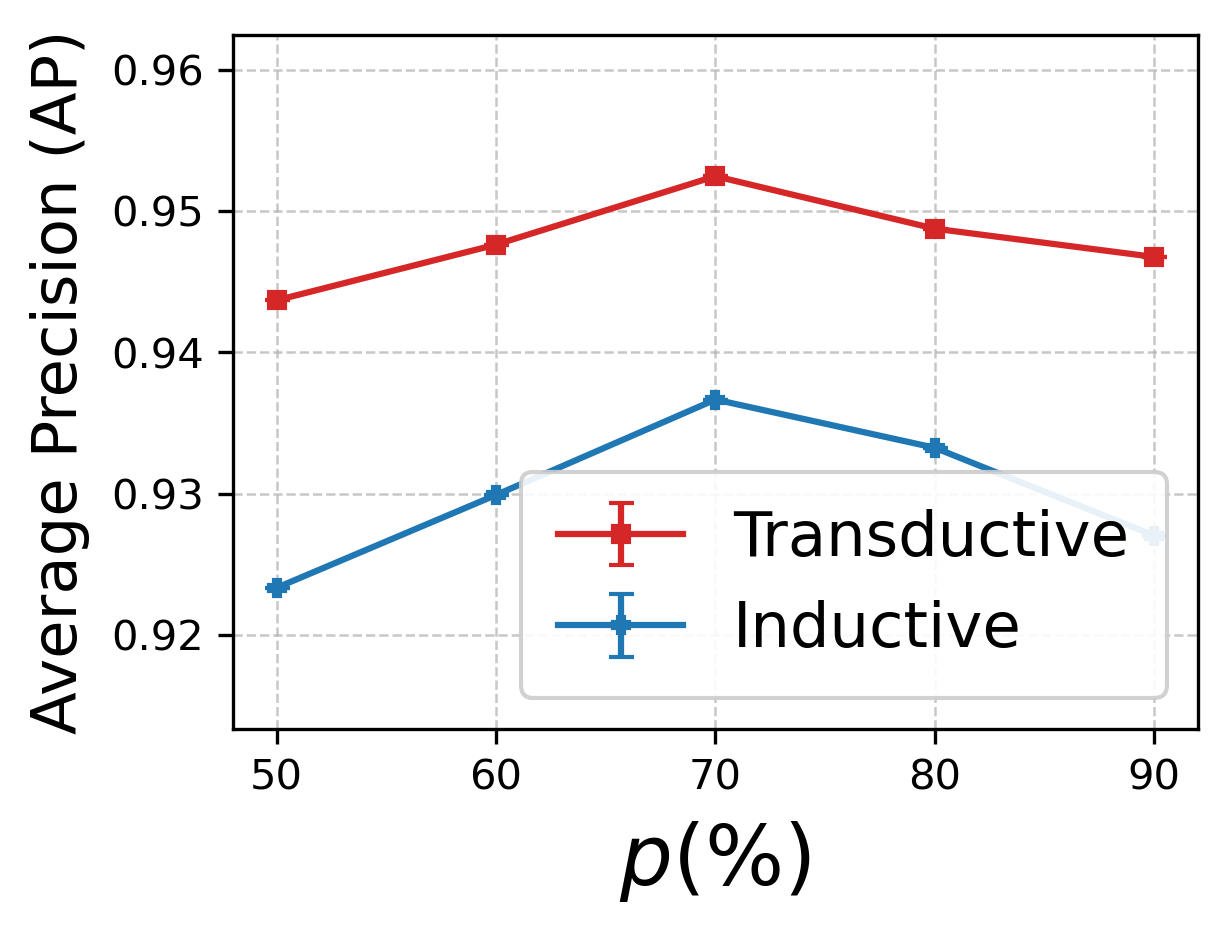}
		\end{subfigure}
		\begin{subfigure}{0.23\textwidth}
			\includegraphics[width=\textwidth]{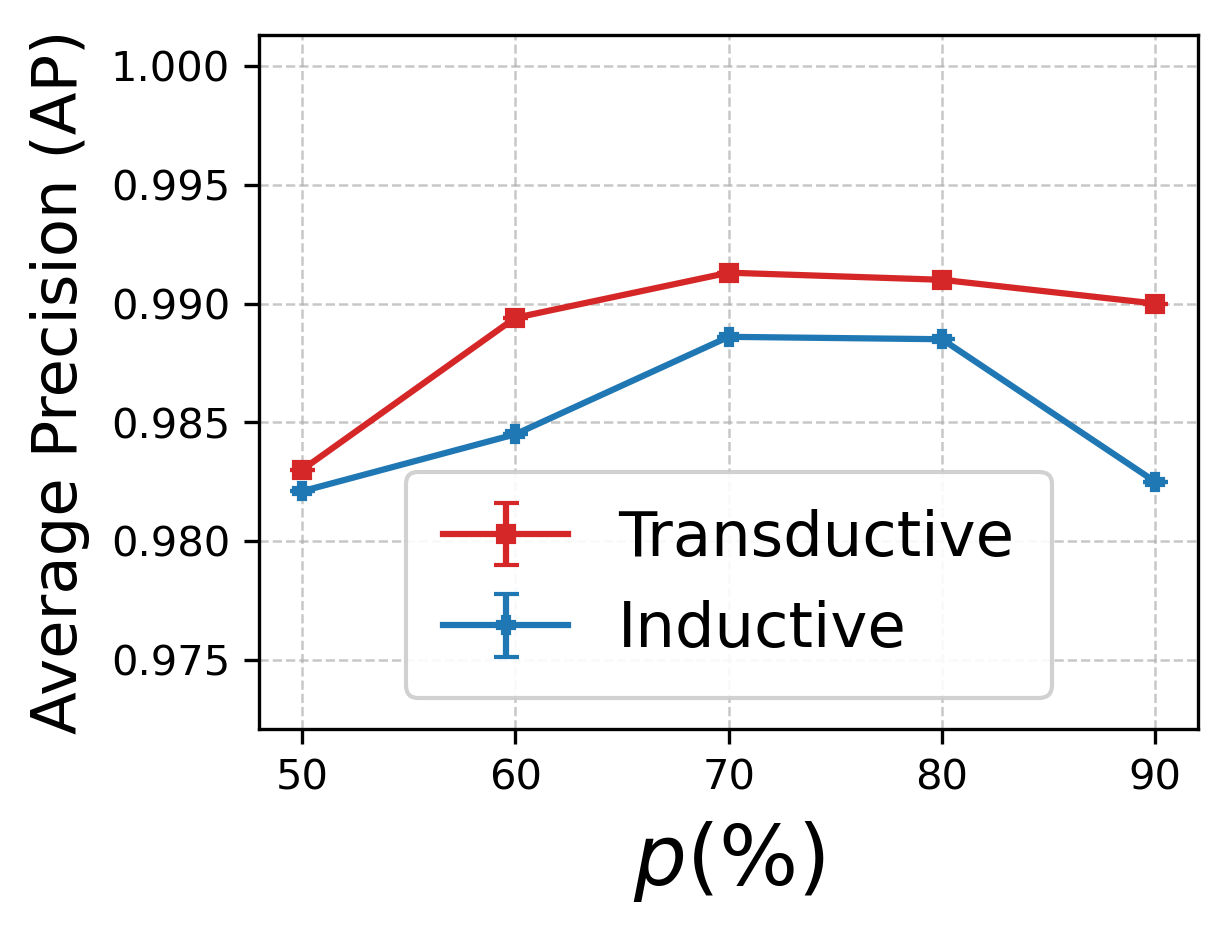}
		\end{subfigure}
		\begin{subfigure}{0.23\textwidth}
			\includegraphics[width=\textwidth]{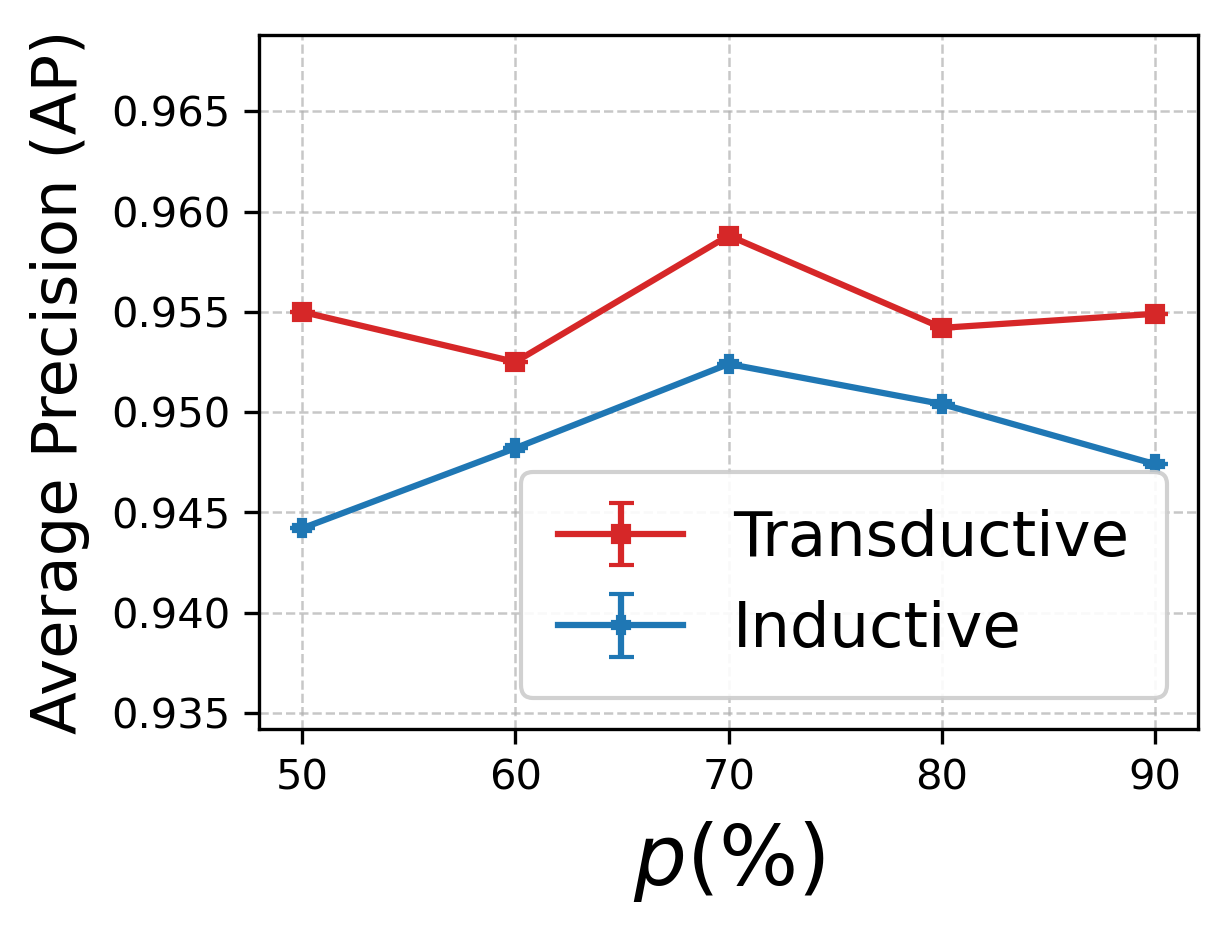}
		\end{subfigure}
		\begin{subfigure}{0.23\textwidth}
			\includegraphics[width=\textwidth]{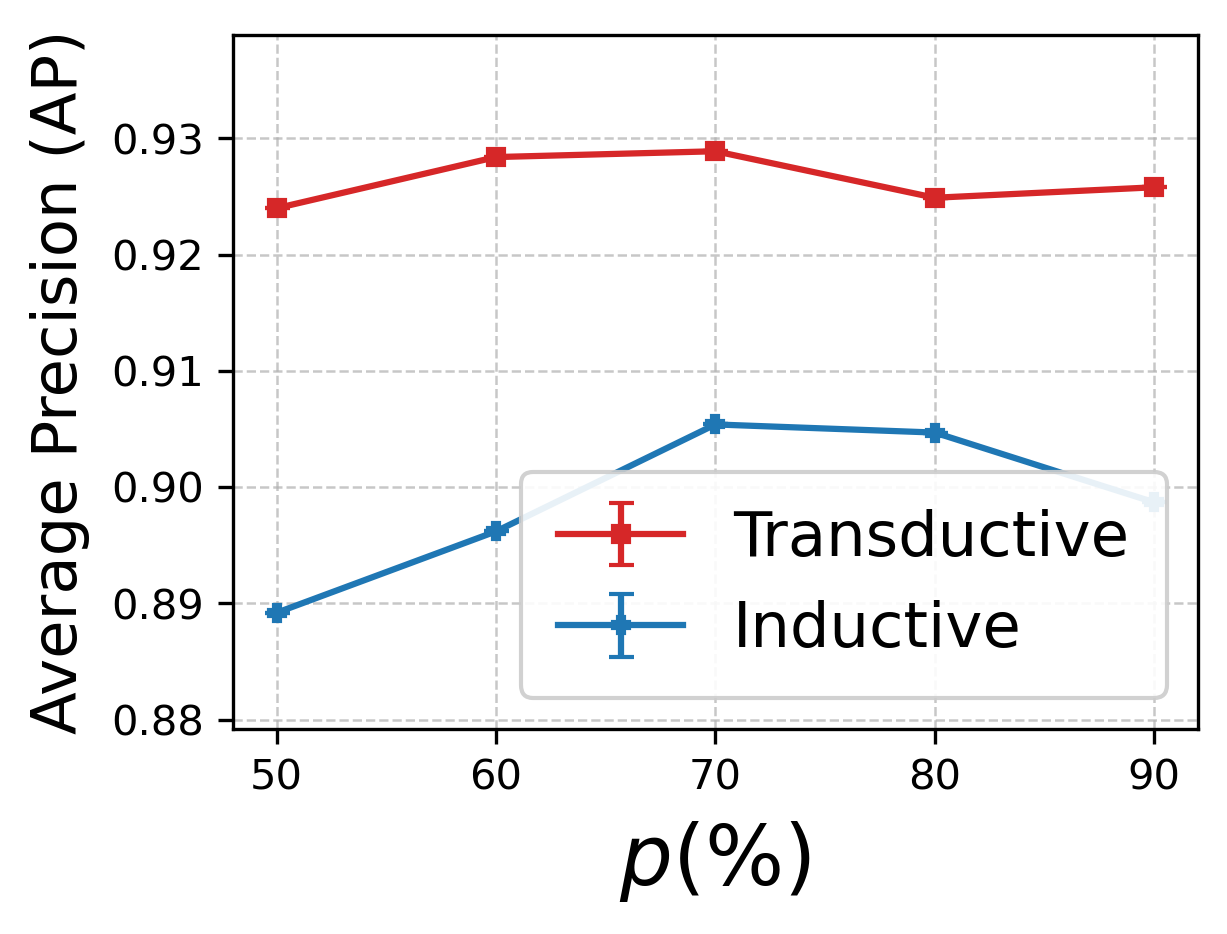}
		\end{subfigure}
		\begin{subfigure}{0.23\textwidth}
			\includegraphics[width=\textwidth]{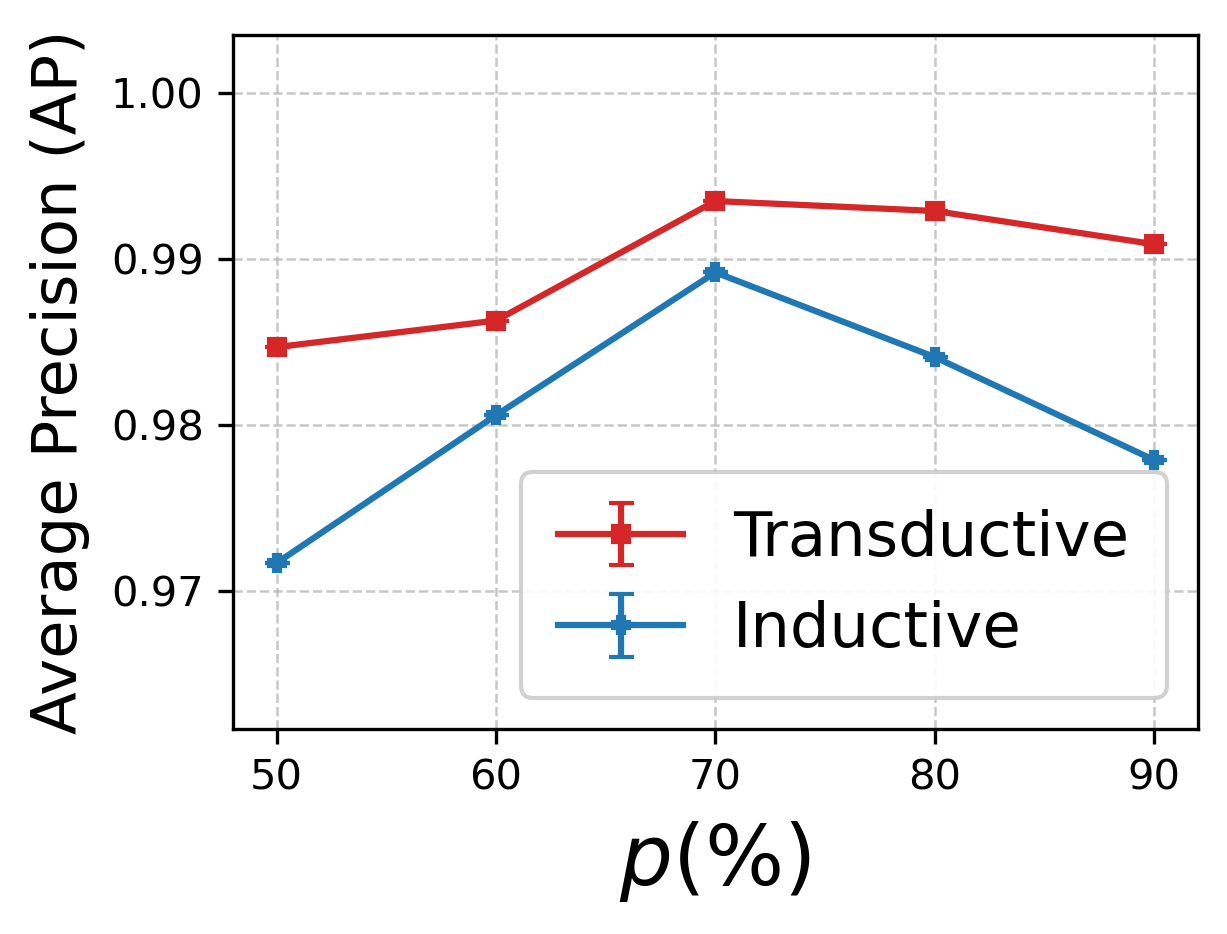}
		\end{subfigure}
		\begin{subfigure}{0.23\textwidth}
			\includegraphics[width=\textwidth]{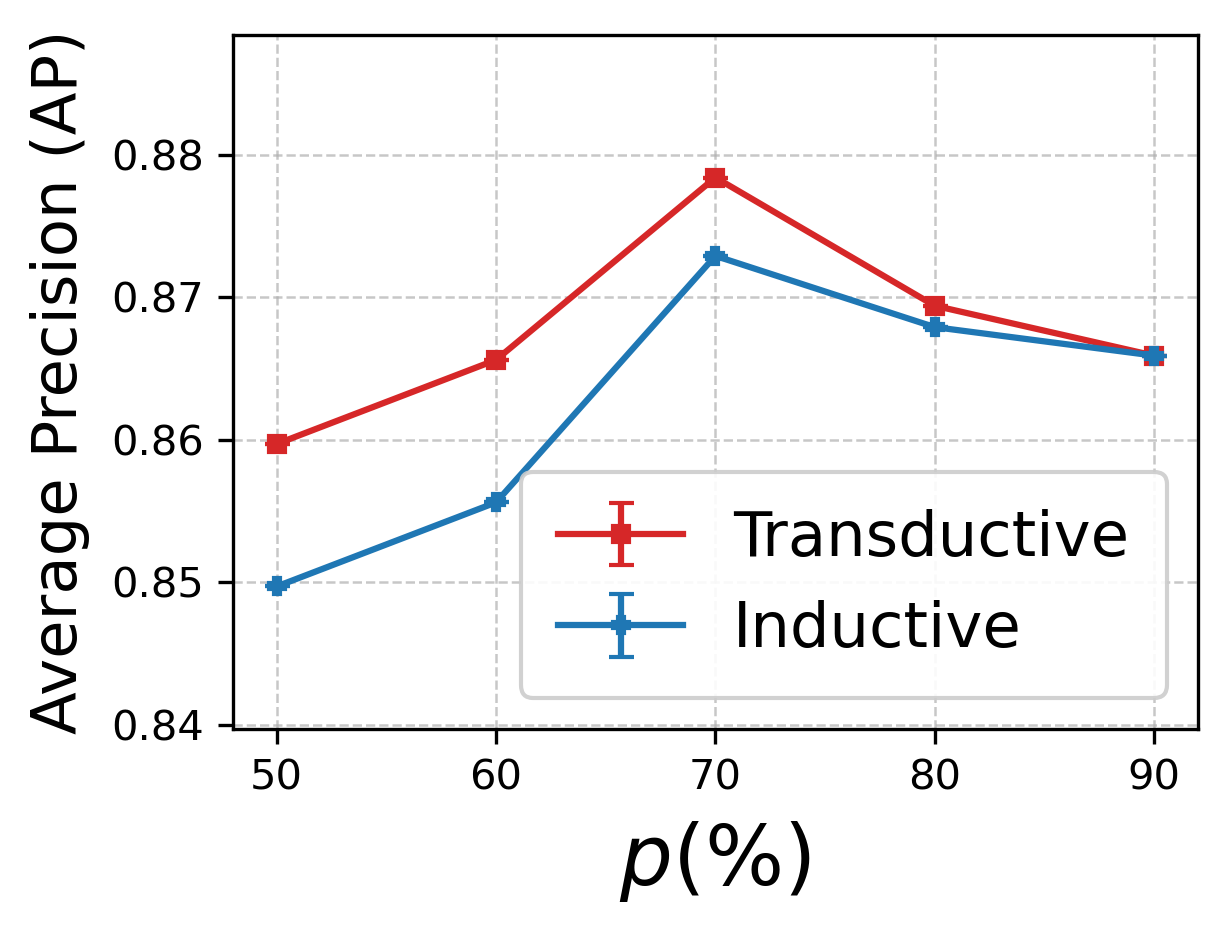}
		\end{subfigure}
		\begin{subfigure}{0.23\textwidth}
			\includegraphics[width=\textwidth]{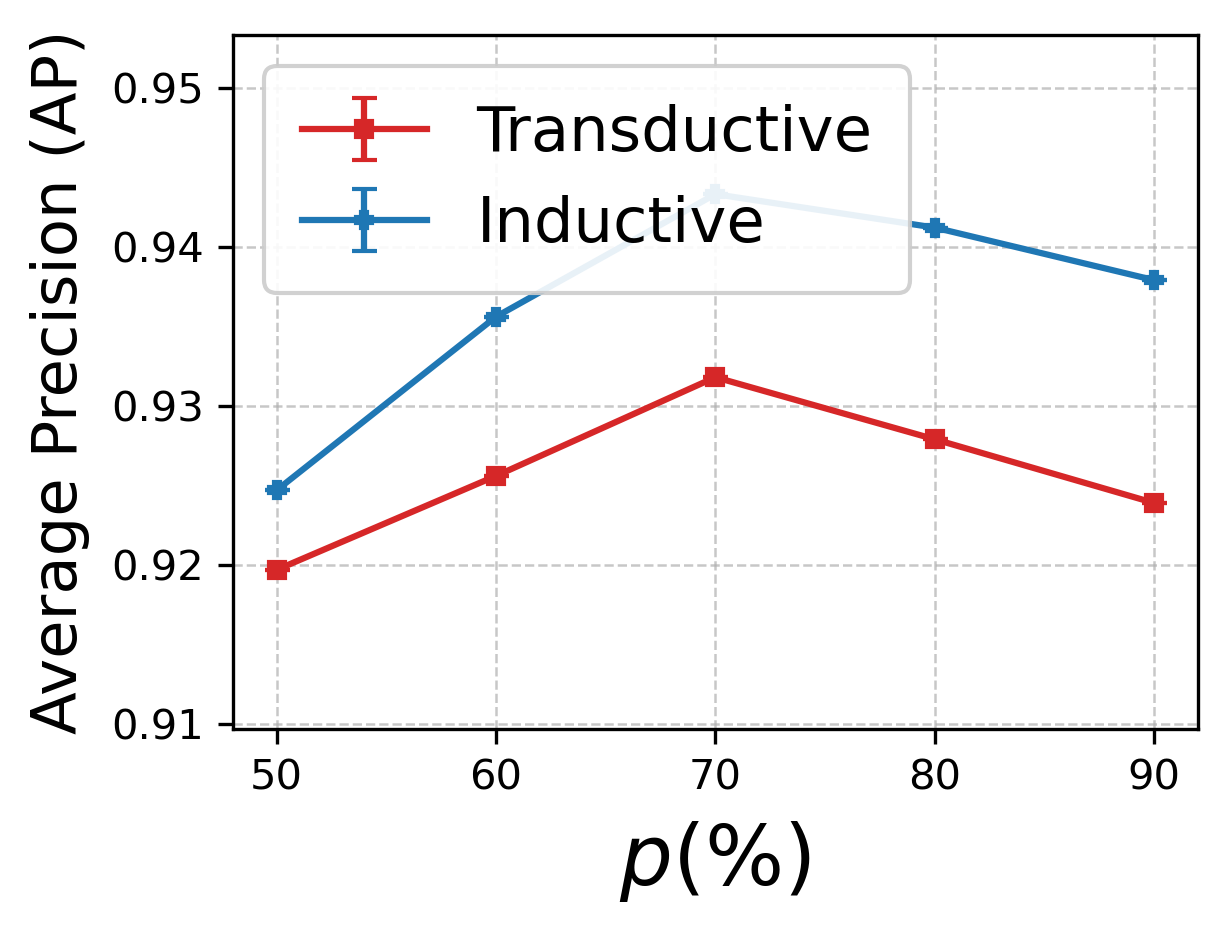}
		\end{subfigure}
		\begin{subfigure}{0.23\textwidth}
			\includegraphics[width=\textwidth]{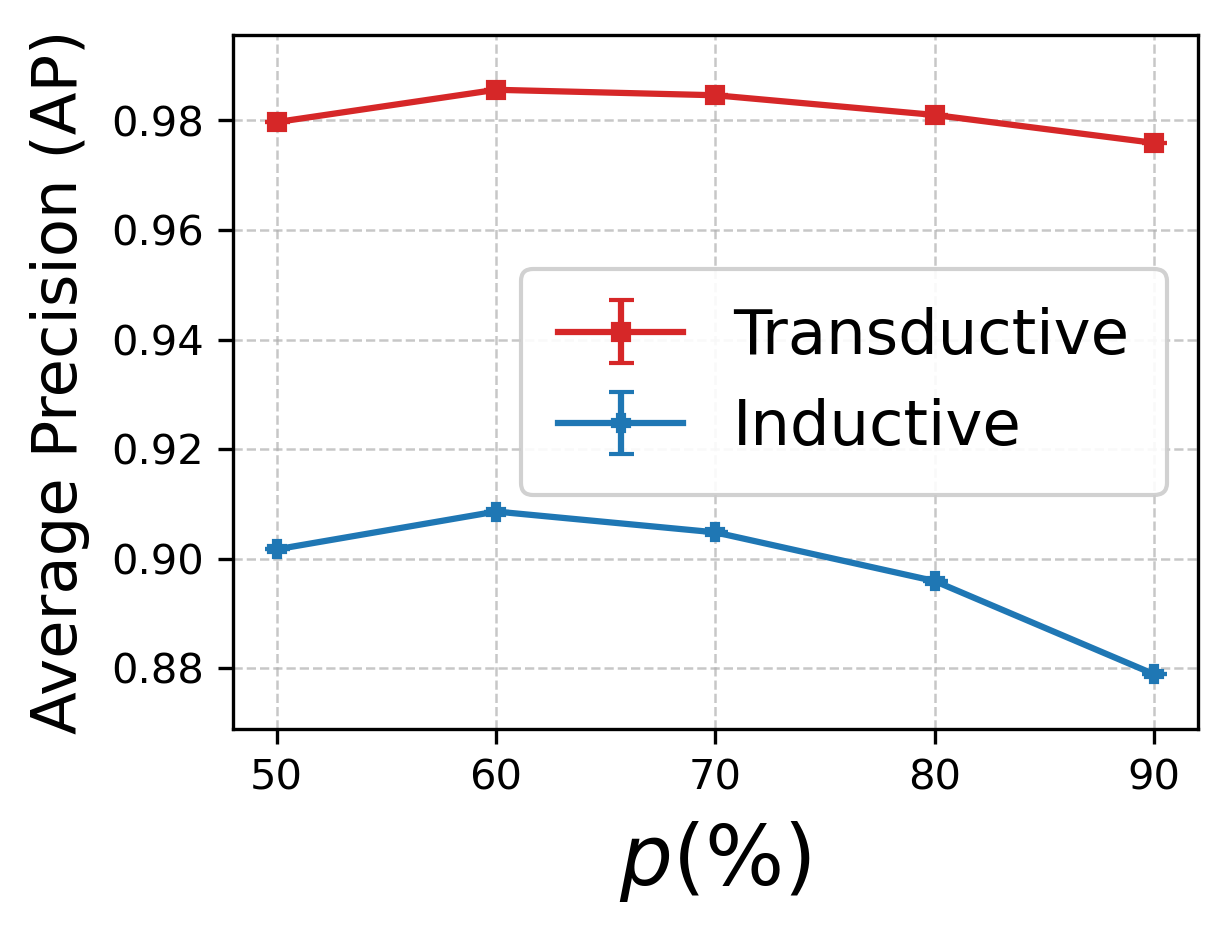}
		\end{subfigure}
		\caption{Results of hyper-parameters sensitivity study in temporal link prediction experiments across different datasets and their average results. $p$: (a) Average, (b) Wikipedia, (c) UCI, (d) Enron, (e) Reddit, (f) Mooc, (g) LastFM, (h) CanParl.}
		\label{fig:hyper1}
	\end{figure}
	\begin{figure}[t]
		\centering
		\begin{subfigure}{0.23\textwidth}
			\includegraphics[width=\textwidth]{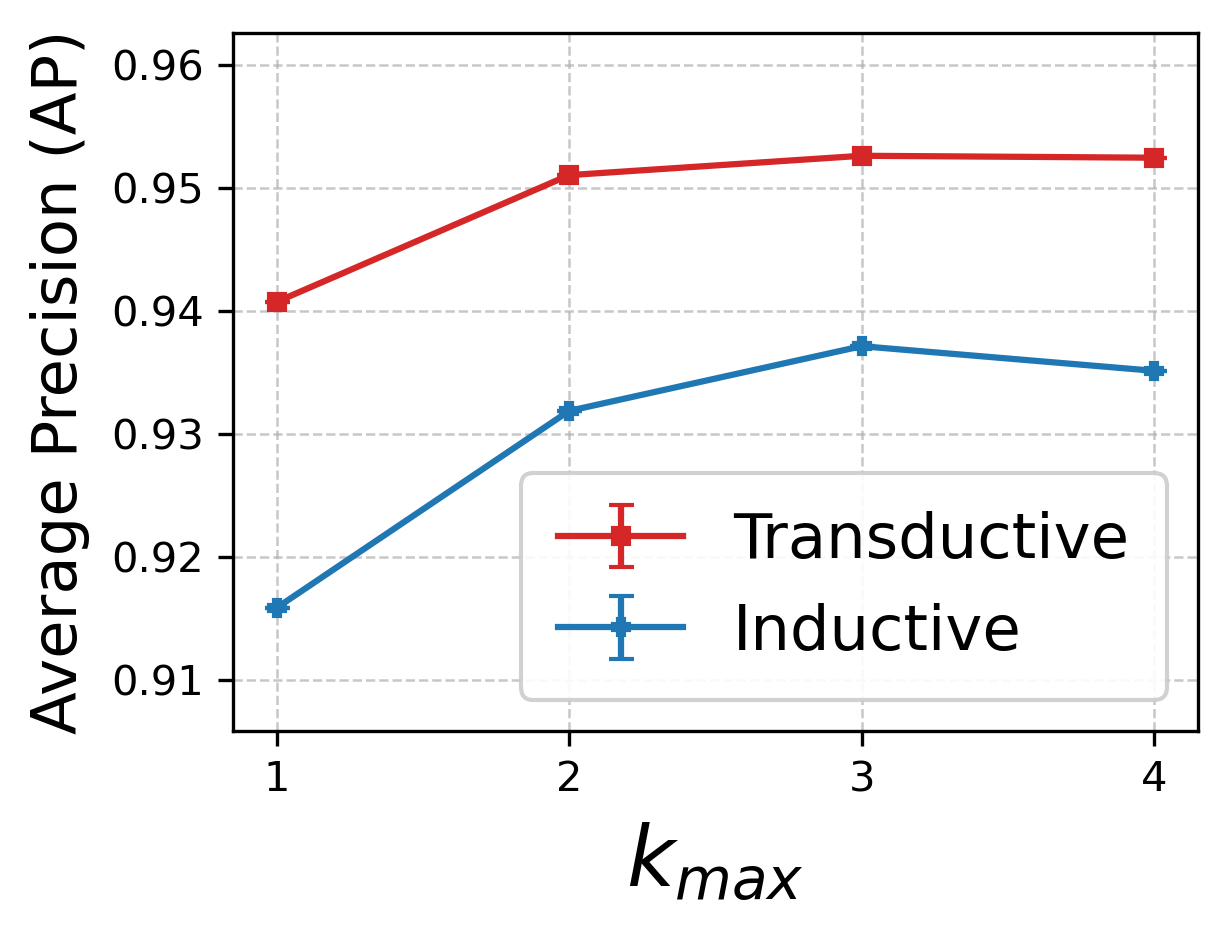}
		\end{subfigure}
		\begin{subfigure}{0.23\textwidth}
			\includegraphics[width=\textwidth]{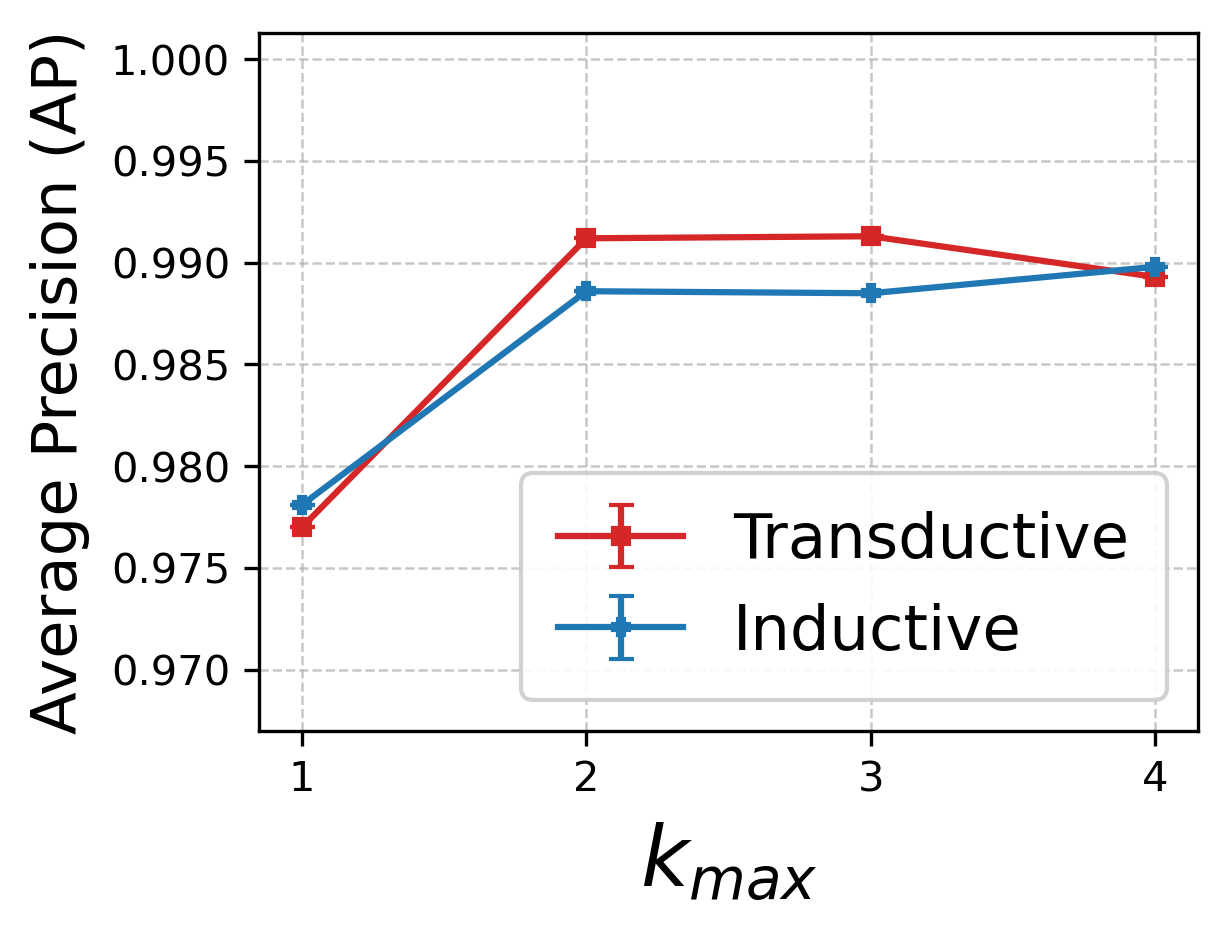}
		\end{subfigure}
		\begin{subfigure}{0.23\textwidth}
			\includegraphics[width=\textwidth]{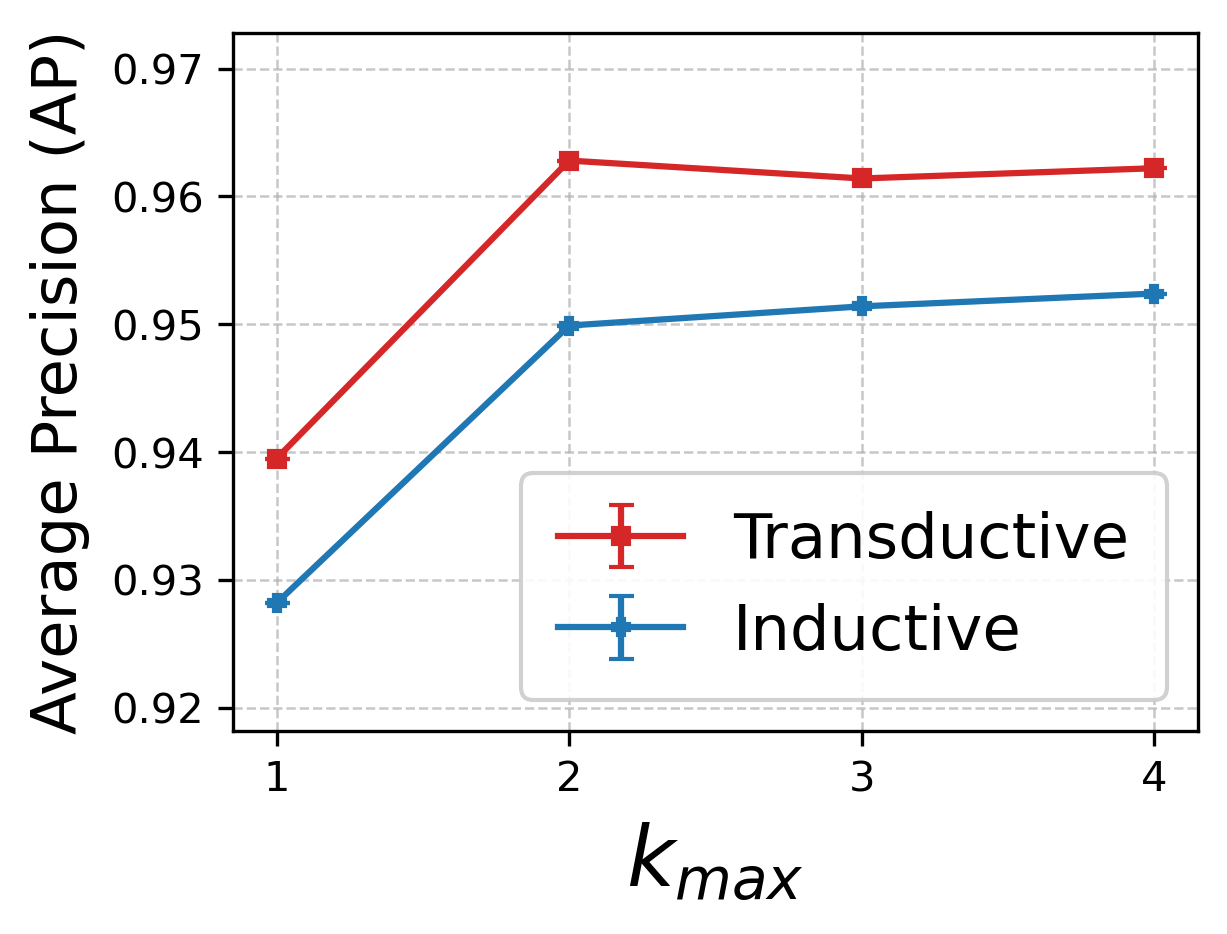}
		\end{subfigure}
		\begin{subfigure}{0.23\textwidth}
			\includegraphics[width=\textwidth]{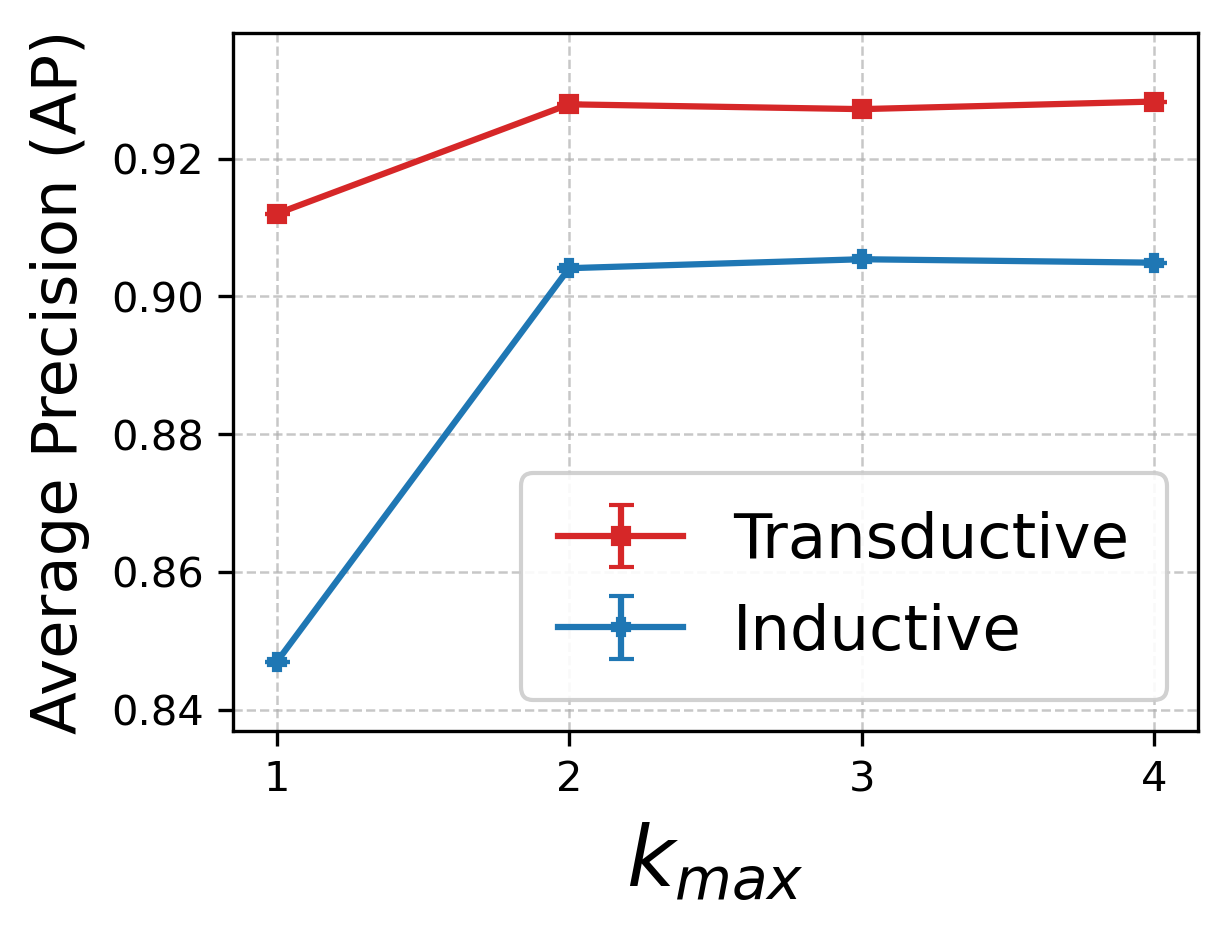}
		\end{subfigure}
		\begin{subfigure}{0.23\textwidth}
			\includegraphics[width=\textwidth]{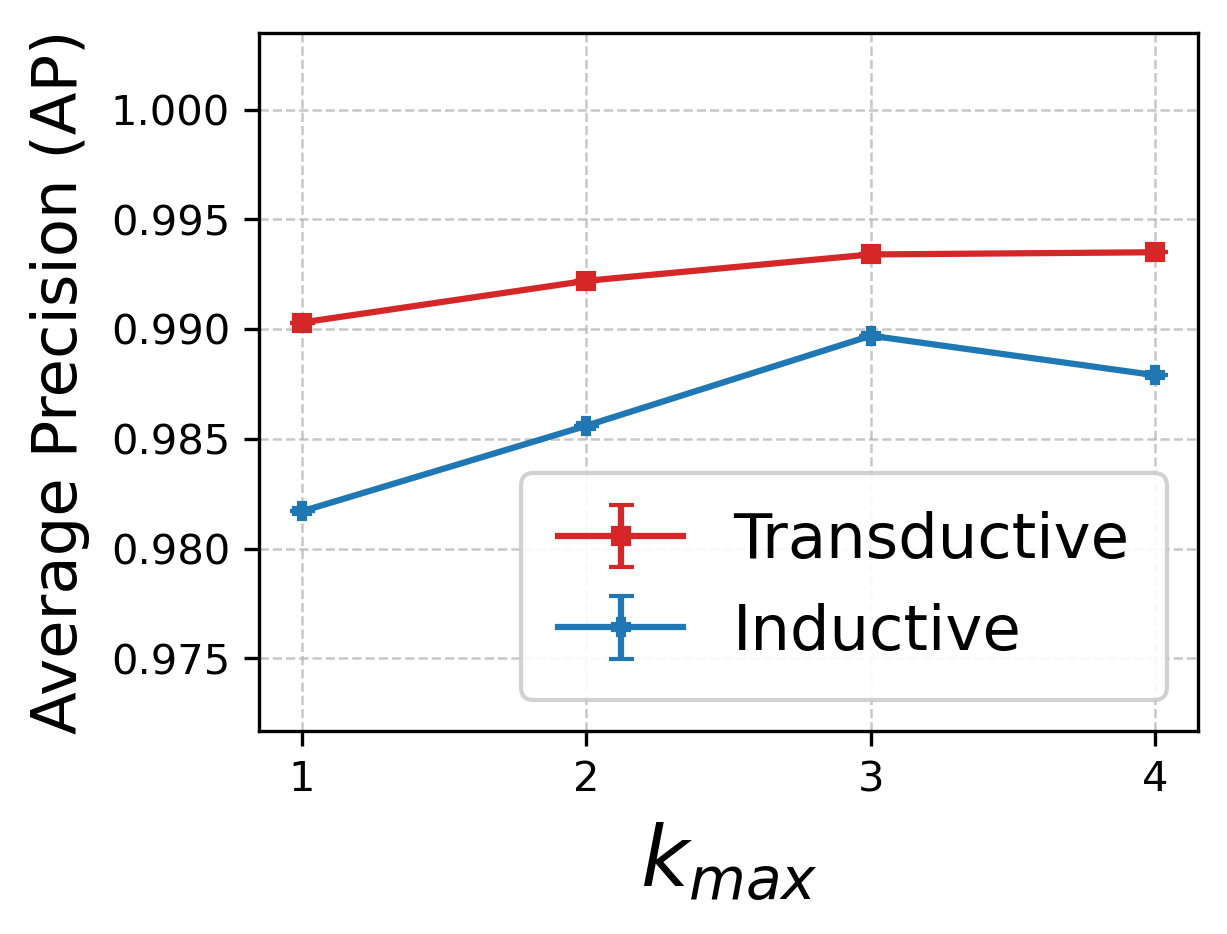}
		\end{subfigure}
		\begin{subfigure}{0.23\textwidth}
			\includegraphics[width=\textwidth]{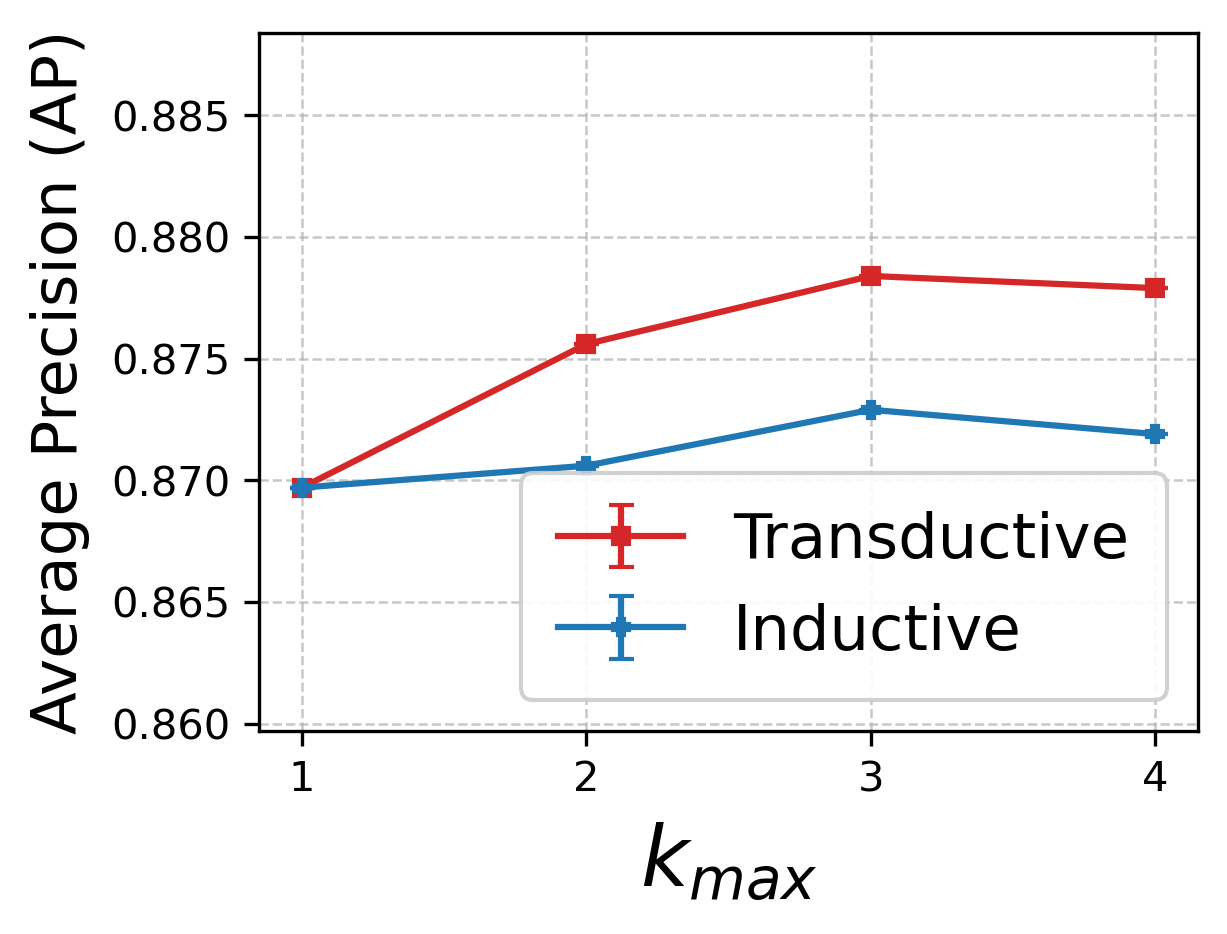}
		\end{subfigure}
		\begin{subfigure}{0.23\textwidth}
			\includegraphics[width=\textwidth]{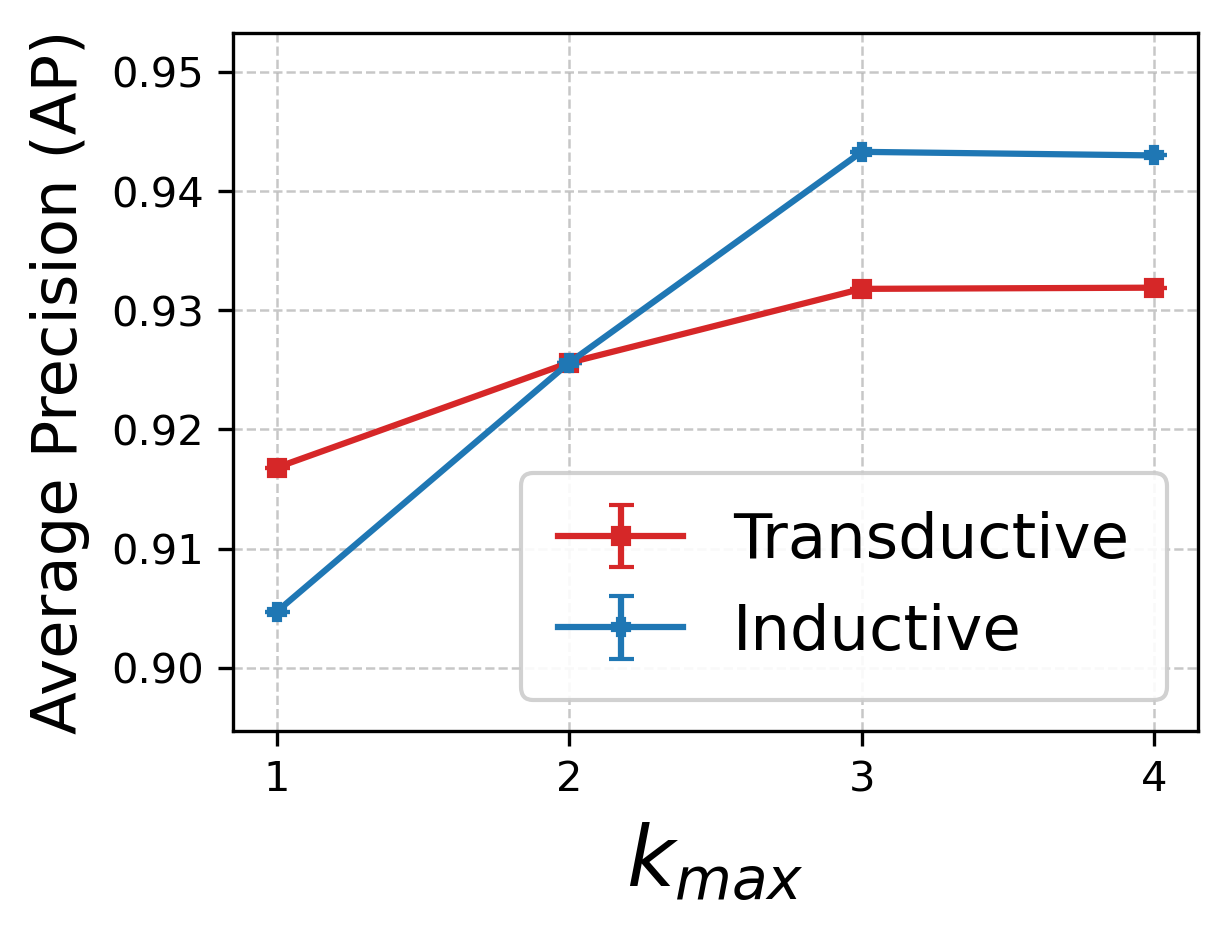}
		\end{subfigure}
		\begin{subfigure}{0.23\textwidth}
			\includegraphics[width=\textwidth]{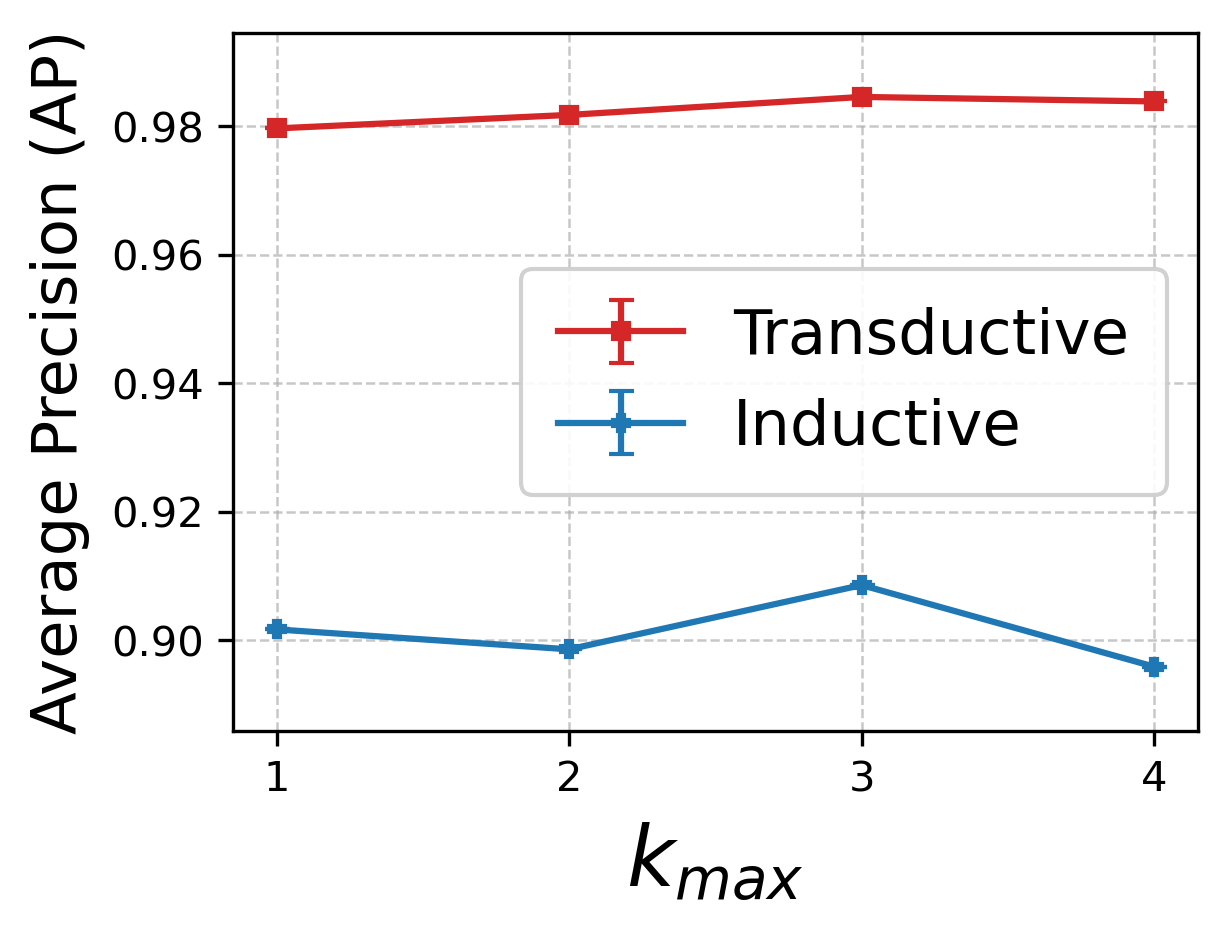}
		\end{subfigure}
		\caption{Results of hyper-parameters sensitivity study in temporal link prediction experiments across different datasets and their average results. $k_{\max}$: (a) Average, (b) Wikipedia, (c) UCI, (d) Enron, (e) Reddit, (f) Mooc, (g) LastFM, (h) CanParl.}
		\label{fig:hyper2}
	\end{figure}
	
	\subsection{Ablation study}\label{sec::abl}
	To test the impact of each module of the model on the final performance, we conducted ablation experiments. Figure \ref{fig:ablation} reports ablation studies on CoDCL. Variants include Full CoDCL and models removing counterfactual learning, temporal encoding, contrastive learning, or similarity constraints.
	
	Counterfactual learning is the dominant contributor. Removing it yields the largest and consistent drops across datasets, exceeding 2\% on UCI and CanParl, indicating that counterfactual reasoning is central to temporal link prediction. Temporal encoding is also critical for temporally complex datasets such as UCI, MOOC, and CanParl, while Wikipedia is comparatively less affected, suggesting weaker reliance on fine-grained temporal order. Contrastive learning provides dataset-dependent gains, with clear benefits on Enron and CanParl. Similarity constraints offer modest but consistent improvements, supporting stability across network types. 
	\begin{table}[h]
		\centering
		\setlength{\tabcolsep}{6pt}
		\caption{Performance comparison of CoDCL with different treatments under both transductive and inductive settings in Wikipedia datasets.}
		\small
		\begin{tabular}{cccc}
			\toprule
			\midrule
			&Method & AP & AUROC \\
			\midrule
			\multirow{7}{*}{Trans}&Dynamic Interaction & {99.13±0.01} & {98.98±0.01} \\
			&Common Neighbors & 98.77±0.01 & 98.88±0.02 \\
			&Degree Similarity & 98.83±0.01 & 98.89±0.01 \\
			&Temporal Proximity & 99.01±0.01 & 98.82±0.01 \\
			&Activity Synchrony & {98.85±0.01} & 98.51±0.02 \\
			&Interaction Frequency & 98.96±0.02 & 98.19±0.01 \\
			&$K$ Core Temporal & 98.83±0.01 & 98.89±0.02 \\
			\midrule
			\multirow{7}{*}{Induc}&Dynamic Interaction & {98.86±0.01} & 98.48±0.01 \\
			&Common Neighbors & 98.50±0.01 & {98.54±0.01} \\
			&Degree Similarity & 98.59±0.02 & 98.48±0.01 \\
			&Temporal Proximity & 98.52±0.01 & 98.52±0.01 \\
			&Activity Synchrony & {98.60±0.01} & 98.51±0.01 \\
			&Interaction Frequency & 98.45±0.02 & 98.36±0.02 \\
			&$K$ Core Temporal & 98.58±0.02 & 98.52±0.01 \\
			\midrule
			\bottomrule
		\end{tabular}
		\label{tab:experimental_results1}
	\end{table}
	\begin{table}[h]
		\centering
		\setlength{\tabcolsep}{6pt}
		\caption{Performance comparison of CoDCL with different treatments under both transductive and inductive settings across UCI datasets.}
		\small
		\begin{tabular}{cccc}
			\toprule
			\midrule
			&Method & AP & AUROC \\
			\midrule
			\multirow{7}{*}{Trans}&Dynamic Interaction & 96.28±0.08 & 94.83±0.10 \\
			&Common Neighbors & 95.87±0.13 & 94.59±0.15 \\
			&Degree Similarity & 95.66±0.21 & 94.27±0.18 \\
			&Temporal Proximity & {96.00±0.12} & {94.73±0.13} \\
			&Activity Synchrony & {95.98±0.15} & {94.76±0.11} \\
			&Interaction Frequency & 95.96±0.12 & 94.19±0.21 \\
			&$K$ Core Temporal & 95.82±0.12 & 94.56±0.14 \\
			\midrule
			\multirow{7}{*}{Induc}&Dynamic Interaction & 95.24±0.08 & {92.87±0.06} \\
			&Common Neighbors & 94.44±0.16 & 92.46±0.13 \\
			&Degree Similarity & 94.33±0.22 & 92.27±0.13 \\
			&Temporal Proximity & {94.58±0.13} & {92.70±0.12} \\
			&Activity Synchrony & {94.61±0.13} & 92.67±0.14 \\
			&Interaction Frequency & 94.55±0.12 & 92.66±0.12 \\
			&$K$ Core Temporal & 94.56±0.14 & 92.61±0.10 \\
			\midrule
			\bottomrule
		\end{tabular}
		\label{tab:experimental_results2}
	\end{table}
	\subsection{Analysis with Different Treatments }
	To assess how treatment selection affects the effectiveness of data augmentation, we conducted sensitivity analyses using six treatments other than dynamic interaction. The aim was to examine performance stability under different causal assumptions and to investigate whether dynamic interaction as a treatment can effectively improve temporal link prediction performance. The alternatives encode distinct dependency views: \textbf{Common Neighbors} measures structural similarity by requiring at least two shared neighbors before the query time; \textbf{Degree Similarity} tests degree homophily via thresholded degree differences; \textbf{Temporal Proximity} captures temporal adjacency through differences in last interaction times; \textbf{Activity Synchrony} checks whether nodes are active within the same time windows; \textbf{Interaction Frequency} summarizes average interaction intensity; \textbf{$K$ Core Temporal} adapts $k$-core structure to temporal settings to match hierarchical positions.
	
	Tables \ref{tab:experimental_results1} and \ref{tab:experimental_results2} show that CoDCL remains effective and robust across datasets and treatments. On Wikipedia, all treatments exceed 98\% performance, and Dynamic Interaction performs best under transductive evaluation. On UCI, results are competitive in the 94--96\% range. For both datasets, transductive performance consistently exceeds inductive performance, with a small gap on Wikipedia and a larger gap on UCI, consistent with differing complexity and the benefit of leveraging test-time structure. Temporal-aware treatment strategies demonstrate superior performance across both datasets. Dynamic Interaction, Temporal Proximity, and Activity Synchrony consistently rank among the top performers, indicating that counterfactual learning effectively exploits temporal signals for improved representations.
	\subsection{Hyperparameter Sensitivity study}
	
	\begin{figure}[t]
		\centering
		\begin{subfigure}{0.34\textwidth}
			\includegraphics[width=\textwidth]{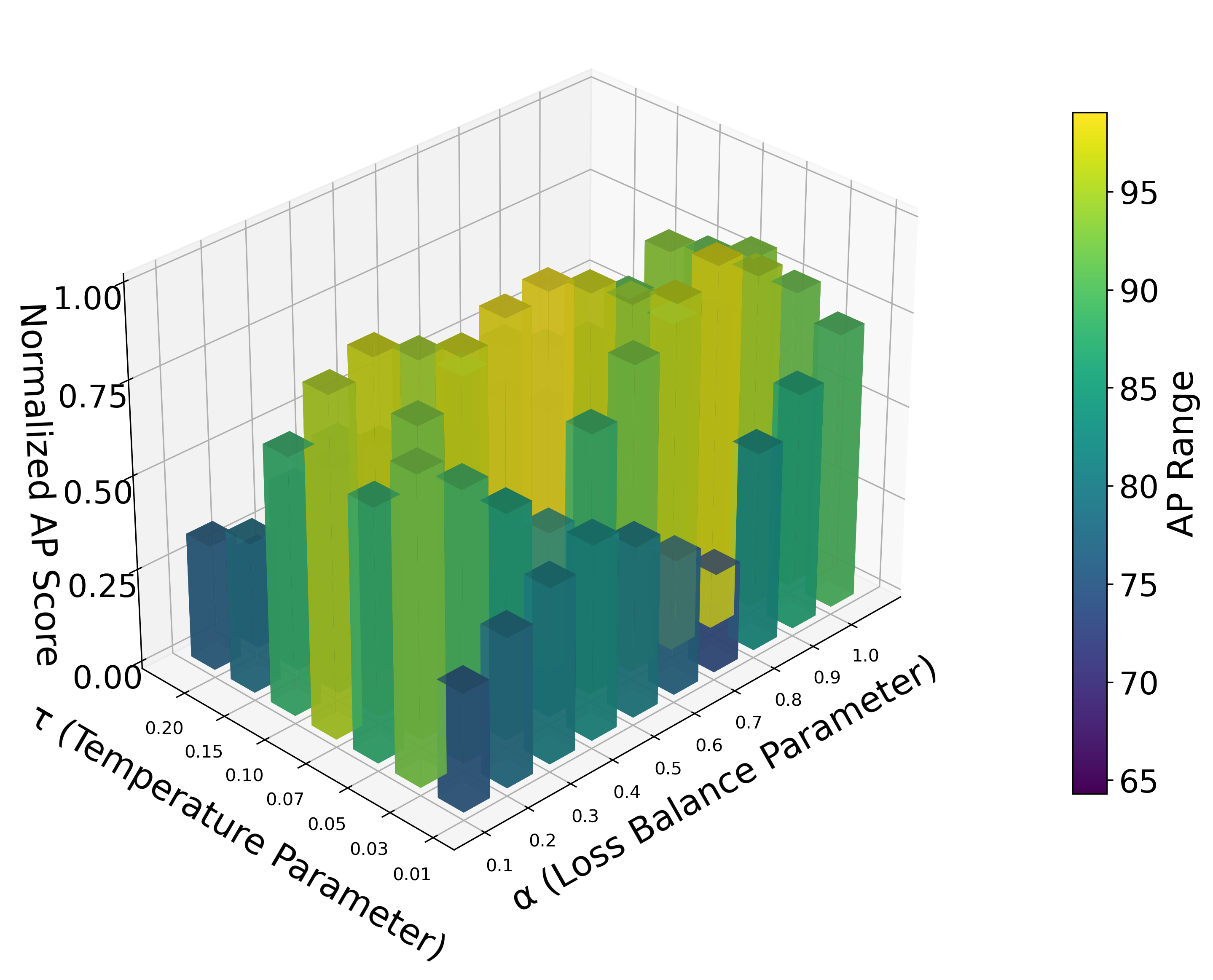}
		\end{subfigure}
		\qquad
		\begin{subfigure}{0.34\textwidth}
			\includegraphics[width=\textwidth]{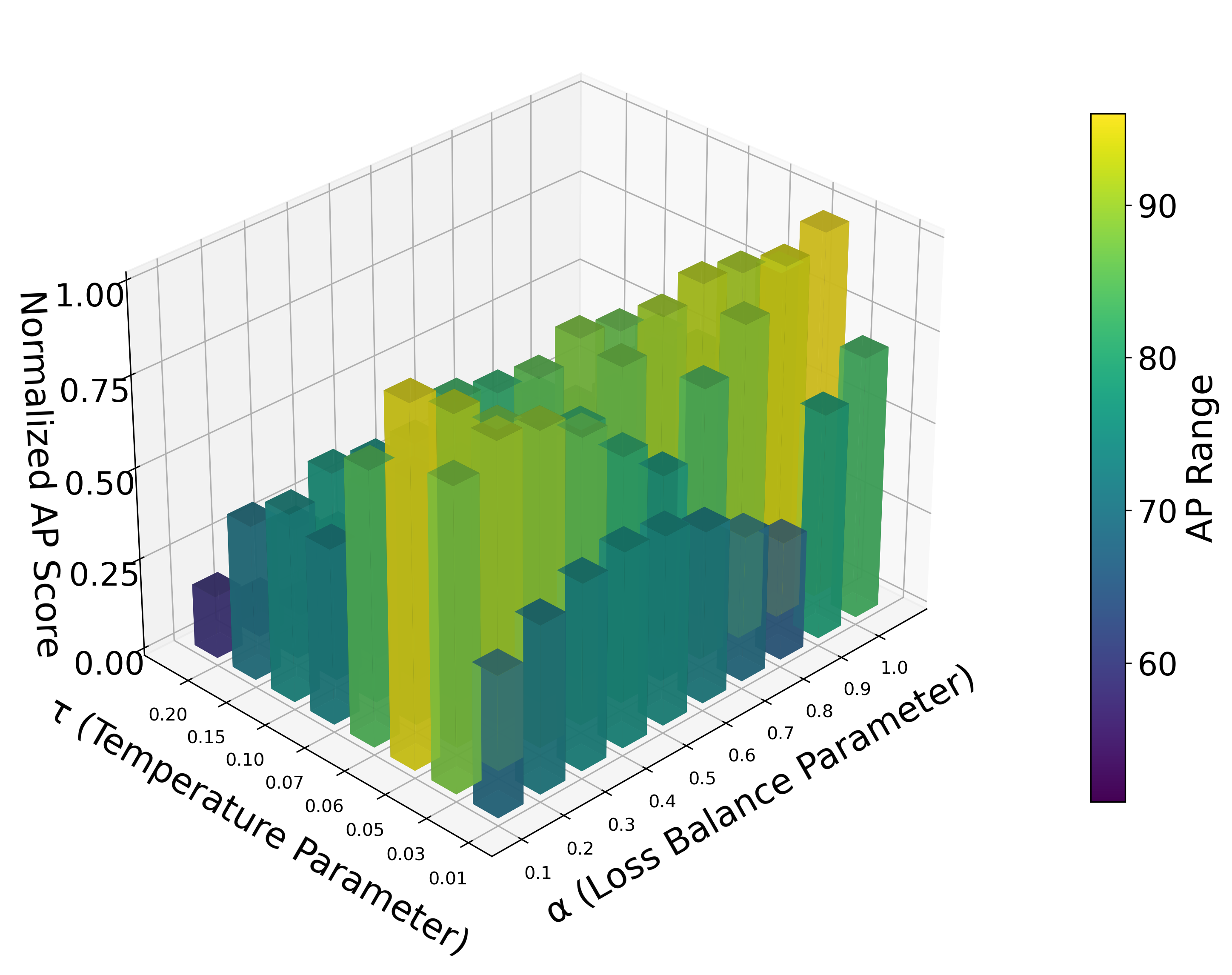}
		\end{subfigure}
		\caption{Performance of CoDCL on different combinations of $\alpha$ and $\tau$: (a) Wikipedia, (b) UCI.}
		\label{fig:wavelet_kernel}
	\end{figure}
	We analyzed the impact of the hyperparameter $p$ on different datasets. As the figure \ref{fig:hyper1} shows, a smaller $p$ leads to too few counterfactual instances, while a larger $p$ introduces a large amount of perturbation noise, both of which reduce prediction accuracy. The Wikipedia and UCI datasets show clearly optimal values, indicating that our proposed counterfactual data augmentation can bring significant performance improvements. The Enron dataset is largely insensitive to $p$, due to its dense communication structure. The Reddit dataset has moderate sensitivity to $p$, with performance relatively stable in the 60\% to 80\% range.
	
	We also examined the impact of the temporal neighborhood size $k_{max}$ on different datasets, as shown in Figure \ref{fig:hyper2}. As $k_{max}$ increases to 2, performance on most datasets gradually improves and tends to saturate, indicating that in counterfactual learning, the benefits of introducing temporal neighborhoods of order two or higher diminish. The performance of Wikipedia stabilized after increasing from $k_{max}=1$ to $k_{max}=2$. UCI showed the strongest dependence on multi-hop neighborhoods, with its performance improvement significantly outperforming single-hop structures. Performance on the Enron dataset gradually improves and quickly saturates, consistent with its dense temporal connectivity.
	
	To analyze the sensitivity of CoDCL to key hyperparameters, we further examined the impact of different combinations of $\alpha$ and $\tau$ on model performance. Figure \ref{fig:wavelet_kernel} shows the performance changes of CoDCL on the Wikipedia and UCI datasets under different combinations of $\alpha$ and $\tau$ hyperparameters, revealing the optimal parameter region for performance improvement. This is because $\alpha$ determines the balance between the basic supervised loss and the counterfactual contrast constraint, while $\tau$ controls the separation strength between similar and dissimilar samples in the representation space. Thanks to co-modeling mechanism of CoDCL, the model can more effectively capture key dynamic patterns under appropriate parameters, thereby improving robustness.

	\section{Conclusion}
	This paper proposes a counterfactual-inspired CoDCL framework for data augmentation to improve temporal link prediction. This framework enhances local dynamic networks based on counterfactual causality and leverages observed and counterfactual temporal patterns through contrastive learning to augment temporal network representations. Extensive experiments on various datasets demonstrate excellent performance in both inductive and transductive settings. Furthermore, it exhibits superior adaptability to complex and dynamic network social scenarios. We also specifically modularize CoDCL, enabling seamless integration into existing temporal network learning frameworks and providing excellent scalability.
	\bibliographystyle{unsrt}
	\bibliography{biblist}
\end{document}